%% file: AnonymousSubmission2027.tex
\documentclass[letterpaper]{article} 

\usepackage{aaai2027}

\usepackage[hyphens]{url}  
\usepackage{graphicx} 
\usepackage{natbib}  
\usepackage{caption} 
\usepackage{booktabs}
\usepackage{makecell}
\usepackage{multirow}

\usepackage[most]{tcolorbox}
\usepackage{enumitem}
\usepackage{capt-of}

\usepackage{xcolor}
\usepackage{tikz}
\usepackage{tikz,graphicx,fontawesome}
\usetikzlibrary{
arrows.meta,
backgrounds,
calc,
fit,
positioning,
shapes.geometric,
shapes.misc
}

\definecolor{cplan}{RGB}{31,97,141}
\definecolor{cexec}{RGB}{22,122,89}
\definecolor{caudit}{RGB}{198,116,20}
\definecolor{cmut}{RGB}{169,50,60}
\definecolor{cmem}{RGB}{104,63,148}

\definecolor{Charcoal}{HTML}{2B2B2B}
\definecolor{CoolGray}{HTML}{7A7D80}
\definecolor{OffWhite}{HTML}{FAF9F7}
\definecolor{LightGray}{HTML}{F1F1F1}
\definecolor{MutedRose}{HTML}{D96C8A}
\definecolor{Blush}{HTML}{F8DDE6}
\definecolor{SageGreen}{HTML}{5F8D77}
\definecolor{PaleSage}{HTML}{DDEBE4}

\definecolor{resultred}{RGB}{220,50,47}
\definecolor{resultorange}{RGB}{230,126,34}

\usepackage{algorithm}
\usepackage{algorithmic}

\title{MANTA: Multi-Agent Network Topology Adaptation for Self-Evolving Multi-Agent Systems}

\author{
MaoXun Huang\textsuperscript{\rm 1,\textdagger},
Jerry Wang\textsuperscript{\rm 2,\textdagger},
Yi-Cheng Lai\textsuperscript{\rm 3,\textdagger}\\
Zhenxing Zhang\textsuperscript{\rm 1},
Claire Cardie\textsuperscript{\rm 1},
Hen-Hsen Huang\textsuperscript{\rm 3}
}

\affiliations{
\textsuperscript{\rm 1}Department of Computer Science, Cornell University, United States\\
\textsuperscript{\rm 2}Department of Computer Science, University of Illinois Urbana-Champaign, United States\\
\textsuperscript{\rm 3}Institute of Information Science, Academia Sinica, Taiwan\\
\textsuperscript{\textdagger}Equal contribution
}

\begin{document}

\maketitle

\begin{abstract}
Large language model-based multi-agent systems improve complex problem solving through task decomposition, agent specialization, information exchange, and intermediate validation. However, existing systems typically treat communication topology as a fixed design choice or an offline optimization target. We introduce \textbf{MANTA}, a framework for \textbf{M}ulti-\textbf{A}gent \textbf{N}etwork \textbf{T}opology \textbf{A}daptation that enables communication structures to self-evolve at inference time. Before execution, MANTA initializes a task-conditioned topology from prior structural experience. During deployment, it monitors collaboration traces and applies bounded structural updates when the current organization becomes insufficient. These updates can modify agent roles, communication links, execution order, information visibility, and validation pathways while preserving the task interface and agent budget. We evaluate MANTA against representative single-agent and multi-agent baselines on five benchmarks spanning information seeking, tool use, planning, workflow execution, and mathematical reasoning. MANTA achieves the highest average score of 74.0, outperforming the strongest baseline by 5.8 percentage points and obtaining the best result on PlanCraft. These results show that inference-time self-improvement can extend to the architecture of collaboration itself.
\end{abstract}

\section{Introduction}

Biological systems adapt not only before encountering their environments, but also through continued interaction with environmental pressures. Modern AI systems are undergoing a similar transition from predominantly offline optimization toward adaptation during execution. Large language models (LLMs) increasingly operate as components of agentic systems that reason, use tools, retain experience, and coordinate with other agents. Rather than relying only on pre-deployment weight updates, these systems can adapt at inference time through prompts, demonstrations, reasoning traces, memories, tools, workflows, and coordination protocols. In-context learning first showed that demonstrations can induce task adaptation during inference \citep{brown2020languagemodelsfewshotlearners}, while chain-of-thought prompting and ReAct demonstrated the value of intermediate reasoning and reasoning-and-acting trajectories \citep{wei2023chainofthoughtpromptingelicitsreasoning, yao2023react}.

Existing self-improvement methods adapt outputs, prompts, reasoning traces,
and memory \citep{madaan2023selfrefineiterativerefinementselffeedback,
shinn2023reflexionlanguageagentsverbal}.
Most approaches, however, adapt individual agents rather than how multiple
agents are organized and communicate.

Multi-agent frameworks decompose complex tasks across specialized roles and collaborative reasoning processes, allowing agents to debate, verify, aggregate, implement, critique, or coordinate toward a shared objective \citep{wu2023autogenenablingnextgenllm,  qian-etal-2024-chatdev, hong2024metagptmetaprogrammingmultiagent, du2023improvingfactualityreasoninglanguage}. Their topology determines who communicates with whom, how information is routed, where validation occurs, and how intermediate outputs are refined.

Topology is still typically optimized at design time rather than adapted during execution. Existing methods jointly optimize prompts and topologies or automate workflow design \citep{zhou2026multiagent, zhang2025aflowautomatingagenticworkflow, hu2025automateddesignagenticsystems, shang2025agentsquareautomaticllmagent}, but generally fix the resulting structure before task execution. Whether topology can self-improve online therefore remains underexplored.

This raises a central question: Can a multi-agent system improve its communication topology while solving each individual task? Unlike prior approaches that rely on training, offline search, or pre-execution optimization, this requires instance-conditioned planning, test-time topology mutation from intermediate collaboration traces, and cross-run transfer of topology experience without updating model weights.

We propose \emph{MANTA}, a framework for \emph{Multi-Agent Network Topology Adaptation} that combines these capabilities. For each task, MANTA plans a topology from accumulated experience, audits the collaboration process, and applies a bounded structural mutation when the current organization becomes insufficient. A cross-run playbook continually distills topology-selection and repair experience across executions.

Across five benchmarks, MANTA achieves the highest average score of $74.0$, outperforming the strongest baseline by $5.8$ points. These results highlight the value of adapting collaboration structures to task needs and execution-time failures. Our contributions are as follows.






\begin{itemize}

\item \textbf{A new problem.} We argue that a multi-agent system's communication structure
should be treated as something the system can change while it works, rather than a design
choice fixed before deployment. We formalize this as \emph{topology-level self-improvement}.

\item \textbf{A method.} We propose MANTA, which organizes a team of agents to fit the task at
hand, watches the collaboration as it unfolds, and reorganizes when the current structure is
clearly failing. What it learns from each task carries over to the next, without any weight
updates or offline search.

\item \textbf{Evidence that it works.} Across five benchmarks, MANTA outperforms single-agent
methods, static multi-agent topologies, and automated workflow-design baselines. Case studies
show it recovering from concrete failures such as an overloaded branch, a missing check, an
agreement reached too early, an action taken twice.

\end{itemize}

\section{Related Work}

\input{pic.tex}
\begin{figure}[t]
\centering
\resizebox{\columnwidth}{!}{
\begin{tikzpicture}[
font=\small,
row/.style={
rounded corners=2pt,
draw=black!35,
line width=0.4pt,
minimum height=0.68cm,
align=left,
inner xsep=4pt,
inner ysep=2pt
},
textrow/.style={row, fill=gray!8},
caprow/.style={row, fill=blue!6, draw=blue!35},
rolerow/.style={row, fill=orange!6, draw=orange!45},
mantarow/.style={
row,
fill=orange!16,
draw=orange!85!black,
line width=0.85pt
},
weightrow/.style={row, fill=purple!7, draw=purple!35},
toparrow/.style={-{Latex[length=2mm]}, line width=0.55pt},
tag/.style={font=\bfseries, align=left},
works/.style={
font=\scriptsize,
align=left,
text=black!65
},
badge/.style={
rounded corners=2pt,
inner xsep=3pt,
inner ysep=1.5pt,
font=\bfseries\scriptsize,
text=white
},
legendbox/.style={
rounded corners=1pt,
draw=black!25,
minimum width=0.24cm,
minimum height=0.15cm,
inner sep=0pt
}
]

\def\xlevel{0.75}
\def\xworks{3.10}
\def\roww{7.55}
\def\workw{5.00}

\node[font=\bfseries\normalsize, align=center] at (4.65,7.70)
{Levels of Self-Improvement};

\draw[toparrow, blue!70!black] (0.05,0.35) -- (0.05,7.10);

\node[
rotate=90,
font=\bfseries\scriptsize,
text=blue!70!black,
anchor=south
] at (0.05,0.55)
{higher level};

\node[
rotate=90,
font=\bfseries\scriptsize,
text=blue!70!black,
anchor=north
] at (0.15,6.90)
{lower level};

\node[textrow, minimum width=\roww cm, anchor=west] at (0.65,0.35) {};
\node[textrow, minimum width=\roww cm, anchor=west] at (0.65,1.20) {};
\node[textrow, minimum width=\roww cm, anchor=west] at (0.65,2.05) {};
\node[caprow, minimum width=\roww cm, anchor=west] at (0.65,2.90) {};
\node[caprow, minimum width=\roww cm, anchor=west] at (0.65,3.75) {};
\node[rolerow, minimum width=\roww cm, anchor=west] at (0.65,4.60) {};
\node[mantarow, minimum width=\roww cm, anchor=west] at (0.65,5.45) {};
\node[weightrow, minimum width=\roww cm, anchor=west] at (0.65,6.30) {};

\node[tag, anchor=west] at (\xlevel,0.35) {L0 Output};
\node[works, anchor=west, text width=\workw cm] at (\xworks,0.35)
{Self-Refine};

\node[tag, anchor=west] at (\xlevel,1.20) {L1 Prompt};
\node[works, anchor=west, text width=\workw cm] at (\xworks,1.20)
{APO, APE, DSPy, MIPRO, PromptBreeder, TextGrad, GEPA};

\node[tag, anchor=west] at (\xlevel,2.05) {L2 Trace};
\node[works, anchor=west, text width=\workw cm] at (\xworks,2.05)
{CoT, Self-Consistency, Zero-shot CoT, ToT, GoT};

\node[tag, anchor=west] at (\xlevel,2.90) {L3 Skill/Tool};
\node[works, anchor=west, text width=\workw cm] at (\xworks,2.90)
{ReAct, Toolformer, Voyager};

\node[tag, anchor=west] at (\xlevel,3.75) {L4 Memory};
\node[works, anchor=west, text width=\workw cm] at (\xworks,3.75)
{Reflexion, DC, Generative Agents, A-MEM, AWM, ACE};

\node[tag, anchor=west] at (\xlevel,4.60) {L5 Agent Role};
\node[works, anchor=west, text width=\workw cm] at (\xworks,4.60)
{CAMEL, ChatDev, MetaGPT, Debate, EoT, MALLM};

\node[
tag,
anchor=west,
text=orange!85!black
] at (\xlevel,5.45)
{L6 Topology};

\node[
works,
anchor=west,
text width=\workw cm,
text=orange!85!black
] at (\xworks,5.45)
{MASS, AFlow, ADAS, AgentSquare, \textbf{MANTA}};

\node[tag, anchor=west] at (\xlevel,6.30) {L7 Weights};
\node[works, anchor=west, text width=\workw cm] at (\xworks,6.30)
{RLHF, SELF};

\node[badge, fill=orange!85!black] at (7.75,5.05)
{our focus};

\draw[
-{Latex[length=1.7mm]},
orange!85!black,
line width=0.45pt
]
(7.35,5.12) -- (6.80,5.40);

\node[legendbox, fill=gray!8] at (1.25,-0.30) {};
\node[anchor=west, font=\scriptsize, text=black!65] at (1.42,-0.30)
{text};

\node[legendbox, fill=blue!6, draw=blue!35] at (2.65,-0.30) {};
\node[anchor=west, font=\scriptsize, text=black!65] at (2.82,-0.30)
{capability};

\node[legendbox, fill=orange!8, draw=orange!45] at (4.55,-0.30) {};
\node[anchor=west, font=\scriptsize, text=black!65] at (4.72,-0.30)
{multi-agent};

\node[legendbox, fill=purple!7, draw=purple!35] at (6.55,-0.30) {};
\node[anchor=west, font=\scriptsize, text=black!65] at (6.72,-0.30)
{weights};

\end{tikzpicture}
}
\caption{\textbf{Levels of self-improvement in agent systems.}
Methods adapt different system components, from outputs and prompts to agent topology and model weights. MANTA self-improves the communication topology.}

\label{fig:self_improvement_pyramid}
\end{figure}
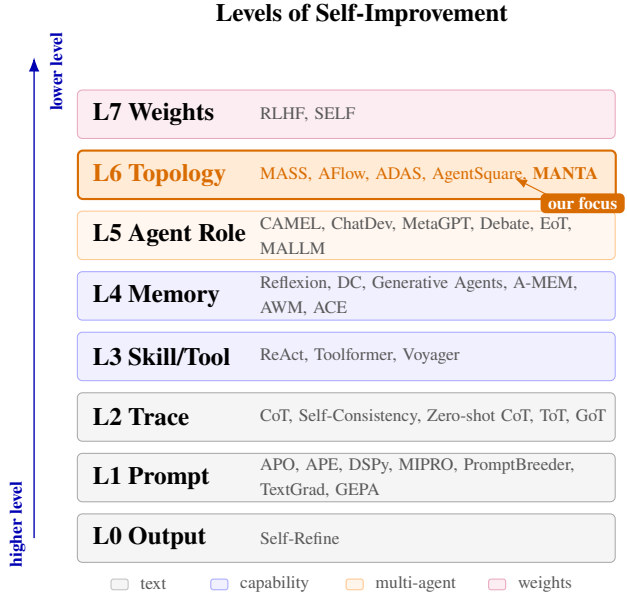

Figure~\ref{fig:self_improvement_pyramid} organizes prior work by the object being adapted, ranging from outputs and prompts to agent capabilities, multi-agent organization, and model weights. This view highlights MANTA's position: whereas most methods improve what an agent produces, observes, remembers, or executes, MANTA improves how multiple agents are organized during task solving.

\textbf{Text-level adaptation.}
Text-level methods adapt outputs, prompts, or reasoning traces. Self-Refine iteratively revises generated answers through feedback \citep{madaan2023selfrefineiterativerefinementselffeedback}. Prompt-level methods search for improved instructions or prompt variants \citep{pryzant2023automaticpromptoptimizationgradient, zhou2023large}, optimize instructions, demonstrations, or modular language-model programs \citep{khattab2023dspycompilingdeclarativelanguage, opsahlong2024optimizinginstructionsdemonstrationsmultistage}, or formulate prompt design as evolutionary, differentiable, or reflective optimization \citep{fernando2023promptbreederselfreferentialselfimprovementprompt, yuksekgonul2024textgradautomaticdifferentiationtext, agrawal2026gepareflectivepromptevolution}. Trace-level methods restructure intermediate reasoning through chain-of-thought, self-consistency, and tree- or graph-based search \citep{wei2023chainofthoughtpromptingelicitsreasoning, wang2023selfconsistencyimproveschainthought, kojima2023largelanguagemodelszeroshot, yao2023treethoughtsdeliberateproblem, besta2024got}. These methods adapt textual or reasoning artifacts while leaving the surrounding agent organization unchanged.

\textbf{Agent-capability adaptation.}
Tool- and skill-oriented methods improve how agents act, call external tools, or acquire reusable behaviors \citep{yao2023react, schick2023toolformerlanguagemodelsteach, wang2023voyageropenendedembodiedagent}. Memory-oriented methods store, retrieve, and refine experience, reflections, or task-solving playbooks across interactions \citep{shinn2023reflexionlanguageagentsverbal, suzgun2025dynamiccheatsheettesttimelearning, Park2023GenerativeAgents, xu2025amemagenticmemoryllm, wang2024agentworkflowmemory, zhang2025agenticcontextengineeringevolving}. These approaches create more persistent and reusable agent state, but generally adapt individual capabilities or context rather than collective organization.

\textbf{Multi-agent organization and topology.}
Role-based systems such as CAMEL, ChatDev, and MetaGPT decompose tasks into specialized agents for communication, planning, coding, reviewing, or management \citep{li2023camelcommunicativeagentsmind, qian-etal-2024-chatdev, hong2024metagptmetaprogrammingmultiagent}. Interaction-based systems coordinate agents through discussion or aggregation, as in multi-agent debate, Exchange-of-Thought, and MALLM \citep{du2023improvingfactualityreasoninglanguage, yin2023exchangeofthoughtenhancinglargelanguage, becker2025mallmmultiagentlargelanguage}. These systems demonstrate the importance of collaboration structure but generally rely on predefined roles and interaction patterns.

More recent work treats multi-agent structure as an optimization target. MASS jointly optimizes prompts and topologies \citep{zhou2026multiagent}, while AFlow, Automated Design of Agentic Systems, and AgentSquare search over agentic workflows or modular system designs \citep{zhang2025aflowautomatingagenticworkflow, hu2025automateddesignagenticsystems, shang2025agentsquareautomaticllmagent}. These approaches establish topology and workflow as meaningful design targets, but primarily optimize them before execution through offline search or aggregate validation performance. The selected workflow is generally fixed once task execution begins.

MANTA instead treats topology as an execution-time object of self-improvement. It uses prior experience to plan an initial task-conditioned organization, then locally repairs communication, coordination, and validation structures when intermediate traces expose a structural deficiency. This distinguishes MANTA from methods that search for a fixed workflow before execution.

\textbf{Weight-level adaptation.}
Weight-level methods improve agents through training or alignment \citep{ouyang2022traininglanguagemodelsfollow, lu2024selfselfevolutionlanguagefeedback}. MANTA instead keeps model weights fixed and adapts the collaboration structure at inference time.

\section{Method}
\input{method.tex}
\section{Experiments}
\input{exps.tex}

\section{Results}
\input{discussion.tex}

\section{Conclusion}

We presented MANTA, a framework that enables multi-agent systems to revise their collaboration topology during task execution. By combining topology planning, trace-based auditing, targeted structural repair, and cross-run experience, MANTA adapts how agents communicate and validate information when the initial organization becomes insufficient. Our findings highlight topology as a distinct layer of self-improvement, where effective adaptation may involve rewiring communication, changing execution order, or adding validation rather than simply increasing the number of agents. This perspective opens a path toward agent systems that can reorganize their collaboration processes in response to emerging failures.

\clearpage
\bibliography{aaai2027}

\clearpage
\appendix
\input{appendix.tex}
\end{document}

%% file: method.tex
\definecolor{cplanning}{RGB}{47,89,166}   
\definecolor{cexecution}{RGB}{34,120,74}  
\definecolor{ccontrol}{RGB}{200,108,18}   
\definecolor{cmemory}{RGB}{112,74,158}    
\definecolor{cglue}{RGB}{95,99,106}       

\definecolor{cplanning}{RGB}{47,89,166}   
\definecolor{cexecution}{RGB}{34,120,74}  
\definecolor{ccontrol}{RGB}{200,108,18}   
\definecolor{cmemory}{RGB}{112,74,158}    
\definecolor{cglue}{RGB}{95,99,106}       

\begin{figure*}[t]
\centering
\tikzset{
  mbox/.style={draw, rounded corners=4pt, align=center, inner xsep=5pt, inner ysep=6pt,
               font=\footnotesize, line width=1pt, fill=white,
               execute at begin node={\hyphenpenalty=10000\exhyphenpenalty=10000\relax}},
  planbox/.style={mbox, fill=cplanning!8, draw=cplanning},
  execbox/.style={mbox, fill=cexecution!8, draw=cexecution},
  ctrlbox/.style={mbox, fill=ccontrol!10, draw=ccontrol},
  membox/.style={mbox, fill=cmemory!8, draw=cmemory},
  gluebox/.style={mbox, fill=cglue!8, draw=cglue},
  iobox/.style={mbox, fill=white, draw=black!70},
  flow/.style={-{Stealth[length=3mm]}, line width=1.1pt, black!70},
  repairflow/.style={flow, cplanning},
  memflow/.style={-{Stealth[length=2.6mm]}, densely dashed, cmemory, line width=1pt},
  lab/.style={font=\footnotesize, fill=white, inner sep=2pt, align=center, text=black!75},
  agentchip/.style={rounded corners=2.5pt, fill=black!72, text=white,
                    font=\fontsize{6}{6}\selectfont\bfseries, inner sep=2.4pt},
  numchip/.style={circle, draw=black!55, fill=white, text=black!70,
                  font=\fontsize{6.5}{6.5}\selectfont\bfseries, inner sep=1.2pt, minimum size=11pt},
  iconbadge/.style={circle, draw=black!55, fill=white, inner sep=1.3pt,
                    minimum size=13.5pt, line width=0.6pt, font=\footnotesize},
  zoneA/.style={rounded corners=10pt, draw=black!22, fill=black!3, line width=0.7pt},
  zoneB/.style={rounded corners=10pt, draw=cmemory!35, fill=cmemory!5, line width=0.7pt},
  ztitle/.style={font=\footnotesize\scshape, text=black!72},
  agent/.style={circle, draw=black!70, fill=white, inner sep=0pt,
                minimum size=3.4mm, line width=0.7pt},
  coord/.style={agent, fill=cplanning!35},
  verif/.style={agent, fill=ccontrol!45},
}

\pgfdeclarelayer{zones}
\pgfsetlayers{zones,background,main}
\resizebox{\textwidth}{!}{%
\begin{tikzpicture}

\begin{scope}[shift={(1.4,1.55)}]
  \fill[cplanning!30, draw=cplanning, rounded corners=1.5pt]
    (0,0) rectangle (0.32,0.24);
  \node[font=\footnotesize, anchor=west, text=black!75]
    at (0.38,0.12) {Planning \& adaptation};

  \fill[cexecution!30, draw=cexecution, rounded corners=1.5pt]
    (3.9,0) rectangle (4.22,0.24);
  \node[font=\footnotesize, anchor=west, text=black!75]
    at (4.28,0.12) {execution};

  \fill[ccontrol!35, draw=ccontrol, rounded corners=1.5pt]
    (6.05,0) rectangle (6.37,0.24);
  \node[font=\footnotesize, anchor=west, text=black!75]
    at (6.43,0.12) {auditing \& control};

  \fill[cmemory!30, draw=cmemory, rounded corners=1.5pt]
    (9.4,0) rectangle (9.72,0.24);
  \node[font=\footnotesize, anchor=west, text=black!75]
    at (9.78,0.12) {memory \& learning};

  \fill[cglue!25, draw=cglue, rounded corners=1.5pt]
    (12.75,0) rectangle (13.07,0.24);
  \node[font=\footnotesize, anchor=west, text=black!75]
    at (13.13,0.12) {deterministic code};

  \node[agentchip] at (16.4,0.12) {Agent};
  \node[font=\footnotesize, anchor=west, text=black!75]
    at (16.8,0.12) {LLM agent};

  \node[numchip] at (18.85,0.12) {1};
  \node[font=\footnotesize, anchor=west, text=black!75]
    at (19.1,0.12) {stage order};
\end{scope}

\node[iobox, anchor=north, text width=0.85cm, minimum height=1.4cm]
  (task) at (0.65,0) {\small\bfseries Task};

\node[planbox, anchor=north, text width=2.35cm] (planner) at (3.15,0)
  {{\small\bfseries Topology Planner}\\[2pt]
   designs a team for the task: topology, agents, roles};

\node[gluebox, anchor=north, text width=2.3cm] (orch) at (6.55,0)
  {{\small\bfseries Orchestrator}\\[2pt]
   validates the topology; spawns the agents};

\node[execbox, anchor=north, text width=2.6cm] (exec) at (10.05,0)
  {{\small\bfseries Turn Executor}\\[2pt]
   runs one collaboration turn};

\node[ctrlbox, anchor=north, text width=2.3cm] (audit) at (14.0,0)
  {{\small\bfseries Trace Auditor}\\[2pt]
   flags process anomalies in the trace};

\node[ctrlbox, anchor=north, text width=1.55cm] (ctrl) at (17.45,0)
  {{\small\bfseries Controller}\\[2pt]
   stop or repair};

\node[execbox, anchor=north, text width=1.95cm] (fin) at (20.4,0)
  {{\small\bfseries Finalizer}\\[2pt]
   votes; synthesizes from evidence};

\node[iobox, anchor=north, text width=1.1cm, minimum height=1.4cm,
      font=\small\bfseries] (ans) at (22.95,0) {Final\\ answer};

\node[coord] (ma0) at (10.05,-2.3) {};
\node[agent] (ma1) at (9.7,-2.9) {};
\node[agent] (ma2) at (10.4,-2.9) {};
\draw[black!55, line width=0.7pt] (ma0)--(ma1) (ma0)--(ma2);

\node[agentchip] (masc) at (9.3,-3.4) {Agent};
\node[font=\footnotesize\itshape, anchor=west, text=cexecution!60!black]
  (mast) at (9.66,-3.4) {task agents};

\begin{scope}[on background layer]
\node[
  draw=cexecution,
  line width=1pt,
  rounded corners=5pt,
  fill=cexecution!6,
  fit=(exec)(ma0)(ma1)(ma2)(masc)(mast),
  inner sep=8pt
] (mas) {};
\end{scope}

\node[
  font=\footnotesize\itshape\bfseries,
  text=cexecution!65!black,
  anchor=south,
  inner sep=2.5pt
] (maslab) at (mas.north) {target multi-agent system};

\node[numchip] at (planner.north west) {1};
\node[iconbadge] at (planner.north east)
  {\textcolor{cplanning}{\faSitemap}};
\node[agentchip] at (planner.south east) {Agent};

\node[numchip] at (orch.north west) {2};
\node[iconbadge] at (orch.north east)
  {\textcolor{cglue}{\faCogs}};

\node[numchip] at (exec.north west) {3};
\node[iconbadge] at (exec.north east)
  {\textcolor{cexecution}{\faPlay}};

\node[numchip] at (audit.north west) {4};
\node[iconbadge] at (audit.north east)
  {\textcolor{ccontrol}{\faSearch}};
\node[agentchip] at (audit.south east) {Agent};

\node[numchip] at (ctrl.north west) {5};
\node[iconbadge] at (ctrl.north east)
  {\textcolor{ccontrol}{\faRandom}};

\node[numchip] at (fin.north west) {7};
\node[iconbadge] at (fin.north east)
  {\textcolor{cexecution}{\faCheckCircle}};
\node[agentchip] at (fin.south east) {Agent};

\coordinate (AX) at (0,-1.0);
\draw[flow] (task.east |- AX) -- (planner.west |- AX);
\draw[flow] (planner.east |- AX) -- (orch.west |- AX);
\draw[flow] (orch.east |- AX) -- (mas.west |- AX);
\draw[flow] (mas.east |- AX) --
  node[lab]{trace}
  (audit.west |- AX);
\draw[flow] (audit.east |- AX) --
  node[lab]{audit\\report}
  (ctrl.west |- AX);
\draw[flow] (ctrl.east |- AX) --
  node[lab]{stop}
  (fin.west |- AX);
\draw[flow] (fin.east |- AX) -- (ans.west |- AX);

\node[font=\footnotesize, anchor=west, text=black!80]
  (rtitle) at (5.45,-4.95)
  {\bfseries Trace-backed repair\normalfont\itshape\ \ (one mutation per run)};

\node[coord] (b0) at (6.0,-5.75) {};
\node[agent] (b1) at (5.65,-6.35) {};
\node[agent] (b2) at (6.35,-6.35) {};
\draw[black!55, line width=0.7pt] (b0)--(b1) (b0)--(b2);

\draw[repairflow] (6.75,-6.05) --
  node[lab, text=cplanning]{mutate}
  (7.75,-6.05);

\node[coord] (c0) at (8.45,-5.75) {};
\node[agent] (c1) at (8.1,-6.35) {};
\node[agent] (c2) at (8.8,-6.35) {};
\node[verif, draw=ccontrol, line width=0.9pt]
  (c3) at (9.2,-5.75) {};

\draw[black!55, line width=0.7pt]
  (c0)--(c1) (c0)--(c2);
\draw[ccontrol, line width=0.9pt, densely dashed]
  (c0)--(c3);

\node[font=\footnotesize, text=ccontrol!85!black, anchor=west]
  (vlab) at (9.45,-5.75) {new verifier};

\node[font=\footnotesize, text=black!70, align=center]
  (rlist) at (8.3,-7.05)
  {edits: add or expand agents, rewire edges,\\
   or adjust visibility};

\begin{scope}[on background layer]
\node[
  draw=cplanning,
  line width=1pt,
  rounded corners=5pt,
  fill=cplanning!6,
  fit=(rtitle)(b0)(b1)(b2)(c0)(c1)(c2)(c3)(vlab)(rlist),
  inner sep=6pt
] (repair) {};
\end{scope}

\node[numchip] at (repair.north west) {6};
\node[iconbadge] at (repair.north east)
  {\textcolor{cplanning}{\faWrench}};

\node[membox, anchor=north, text width=3.2cm] (stm) at (15.0,-4.55)
  {{\small\bfseries Short-term playbook}\\
   {\itshape (this run)}\\[2pt]
   each turn's topology, findings, and decision};

\node[iconbadge] at (stm.north east)
  {\textcolor{cmemory}{\faClipboard}};

\draw[repairflow, rounded corners=4pt]
  (ctrl.south) --
  (ctrl.south |- 0,-7.0) --
  (repair.east |- 0,-7.0);

\node[lab, anchor=west, text=cplanning]
  at ([xshift=3pt]ctrl.south |- 0,-3.7)
  {repair needed};

\draw[repairflow, rounded corners=4pt]
  (repair.west) --
  (3.5,-5.75 |- repair.west) --
  ([xshift=10pt]planner.south);

\node[lab, anchor=west, text=cplanning]
  at (3.7,-3.6)
  {a final turn runs on\\ the revised topology};

\draw[memflow]
  (audit.south) --
  node[lab, pos=0.5]{records\\each turn}
  (stm.north);

\draw[memflow]
  (stm.west) --
  node[lab, pos=0.5]{informs}
  (repair.east |- stm.west);

\node[membox, anchor=north, text width=4.1cm] (ltm) at (2.6,-8.6)
  {{\small\bfseries Long-term playbook}\\[2pt]
   standing principles \,$\cdot$\, how to choose a topology
   \,$\cdot$\, lessons from experience};

\node[iconbadge] at (ltm.north east)
  {\textcolor{cmemory}{\faBook}};

\node[membox, anchor=north, text width=2.6cm] (refl) at (20.4,-8.6)
  {{\small\bfseries Reflector}\\[2pt]
   periodically rewrites the lessons};

\node[numchip] at (refl.north west) {8};
\node[iconbadge] at (refl.north east)
  {\textcolor{cmemory}{\faRefresh}};
\node[agentchip] at (refl.south east) {Agent};

\draw[memflow]
  ([xshift=-15pt]planner.south |- ltm.north) --
  node[lab, pos=0.3]{read at initial plan\\and repair time}
  ([xshift=-15pt]planner.south);

\draw[memflow]
  (fin.south) --
  node[lab, pos=0.55]{run outcome\\(process signals)}
  (fin.south |- refl.north);

\draw[memflow]
  (refl.west |- 0,-9.4) --
  node[lab, pos=0.5]{rewrites the lessons (batched)}
  (ltm.east |- 0,-9.4);

\begin{pgfonlayer}{zones}
\node[
  zoneA,
  fit=(task)(planner)(orch)(mas)(maslab)(audit)(ctrl)(fin)(ans)(repair)(stm),
  inner sep=10pt
] (z1) {};

\coordinate (z2left) at (z1.west |- ltm.west);
\coordinate (z2right) at (z1.east |- refl.east);

\node[
  zoneB,
  fit=(ltm)(refl)(z2left)(z2right),
  inner xsep=0pt,
  inner ysep=6pt
] (z2) {};
\end{pgfonlayer}

\node[ztitle, rotate=90, anchor=south]
  at (z1.west) {per-task loop};

\node[ztitle, rotate=90, anchor=south]
  at (z2.west) {cross-run memory};

\end{tikzpicture}%
}

\caption{\textbf{MANTA overview.} MANTA first plans a task-conditioned
multi-agent topology and executes the task under that structure. The Trace
Auditor flags process-observable anomalies in the collaboration trace; it
does not judge answer correctness. The Controller either finalizes the
answer or returns the diagnosis for one bounded topology revision. A
short-term playbook supports repair within the current run, while a
long-term playbook transfers reusable topology-selection lessons across
runs.}
\label{fig:manta}
\end{figure*}
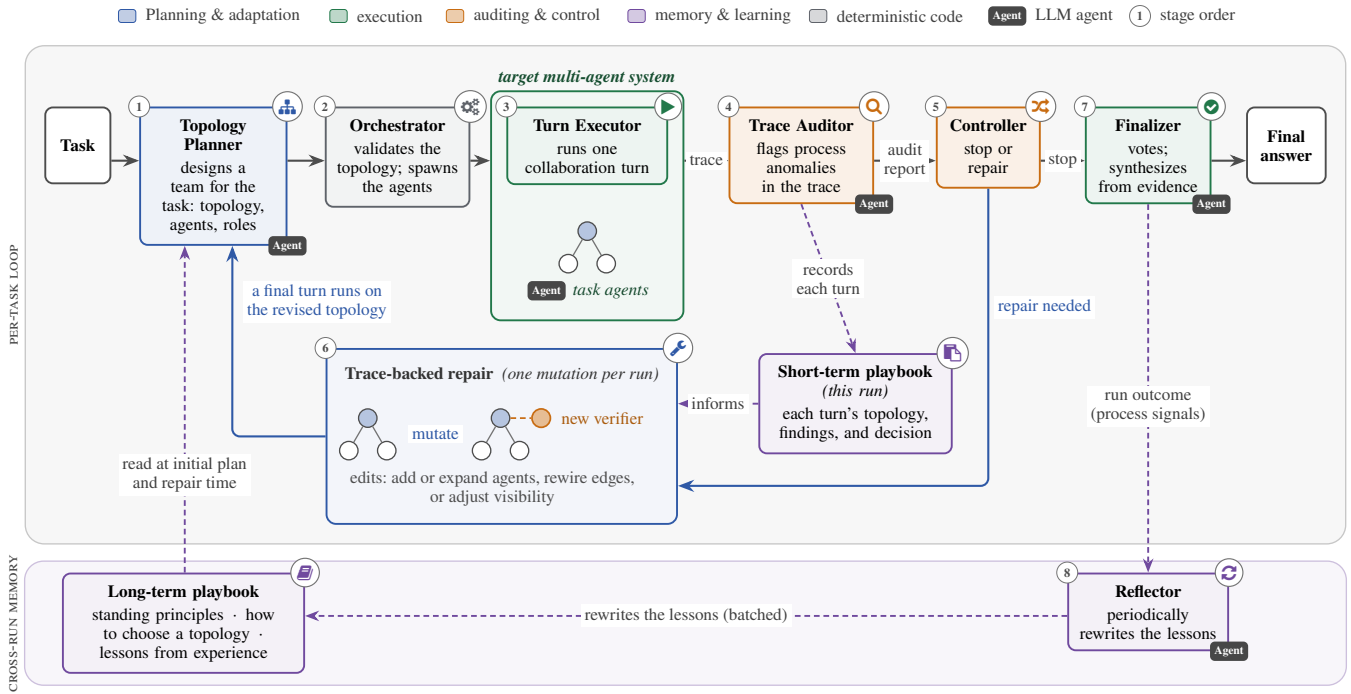

MANTA chooses a communication topology for each task and can revise it
during execution. This section defines its topology representation,
orchestration loop, and memory.

\subsection{System Overview}
\label{sec:method-overview}

Figure~\ref{fig:manta} shows two nested layers. The \textbf{target
multi-agent system} solves the task through reasoning, tool calls, and
message exchange. The \textbf{orchestration layer} selects the agents,
communication links, and information visibility without solving the task.
It combines three LLM components called the \emph{Topology Planner},
\emph{Trace Auditor}, and \emph{Skill Reflector} with deterministic code
that validates topologies, runs agents, routes messages, controls repair,
and decides when to stop.

For each task, the Planner chooses the team size, interaction pattern, and
roles. After one collaboration turn, the Auditor reports observable process
risks rather than answer correctness. The Controller then finalizes the
answer or permits one bounded repair and a final turn.

Two playbooks provide memory. One records topology and audit history within
the current run. The other stores lessons across runs and is updated by the
Skill Reflector every $N$ runs. It uses only process signals and never
receives the benchmark verdict.

\providecommand{\mantaicon}{%
\raisebox{0.05ex}{%
\begin{tikzpicture}[scale=0.12,baseline={(current bounding box.center)}]
\fill[black]
(0,0)
.. controls (-1.8,0.9) and (-3.2,0.4) .. (-4.2,-0.4)
.. controls (-2.8,-0.2) and (-1.8,-0.8) .. (-0.6,-1.1)
.. controls (-0.25,-1.55) and (0.25,-1.55) .. (0.6,-1.1)
.. controls (1.8,-0.8) and (2.8,-0.2) .. (4.2,-0.4)
.. controls (3.2,0.4) and (1.8,0.9) .. (0,0)
-- cycle;
\draw[black,line width=0.7pt]
(0,-1.05)
.. controls (0,-1.65) and (0.15,-2.05) .. (0.45,-2.45);
\fill[white] (-0.9,-0.25) circle (0.16);
\fill[white] (0.9,-0.25) circle (0.16);
\end{tikzpicture}%
}%
}

\subsection{Topology Representation}
\label{sec:method-spec}

A topology specification defines a versioned and validated multi-agent
system. Each agent has a \emph{structural role}, such as coordinator,
worker, verifier, debater, or voter, that determines its graph position. It
also has a \emph{stage role} as a worker or critic. Workers produce
contributions supported by evidence. Critics check and revise visible
claims. An agent may also receive a domain persona and a list of allowed
tools.

Agents belong to \emph{pattern groups} with predefined interaction
structures. Groups can be nested, which lets an agent delegate work to a
subgroup and synthesize its reports. Direct connections may supplement this
tree. Each context policy controls the messages, senders, and shared
evidence an agent can access, how far its messages travel, and whether it
receives full or summarized reports. Code enforces these policies whenever
information is read.

Before execution, MANTA checks group membership, attachment cycles, roles,
and the agent budget. Invalid proposals are rejected. The representation
includes a single agent and the six fixed workflows as special cases while
also supporting nested combinations.

\subsection{The Orchestration Loop}
\label{sec:method-loop}

\paragraph{Query-conditioned planning.}
The Planner receives the task and the experience memory without access to the benchmark identity or a hand-designed topology. It analyzes the task requirements and likely process risks, then produces a compact plan specifying the interaction pattern, agent count, and any optional verifier or nested groups. Deterministic code expands this plan into a complete topology, assigns roles and context policies, and validates all structural constraints. 

\paragraph{Turn execution.}
Each collaboration turn executes the topology recursively according to its group structure. Agents communicate through structured relay packets rather than raw transcripts. These packets summarize their answers, evidence, confidence, and unresolved issues. A shared-context controller enforces the visibility policy of every agent. During execution, identical tool calls are deduplicated, while an append-only evidence ledger preserves the claims and supporting evidence produced across branches.

\paragraph{Trace auditing and repair.}
After each turn, an LLM Auditor scans structured artifacts, tool records, relay
packets, confidence, unresolved issues, and evidence visibility. It may also add a new trace-grounded process flag, but it cannot consult the
benchmark answer or verdict. We call a run \emph{flagged} when the initial audit recommends
repair because at least one repairable flag has medium or high severity. A
\emph{clean} run does not activate this gate. These terms describe the
observed collaboration process rather than answer correctness.

For a general task, the Planner returns one mutation with at most
three operations. It may add or expand an agent, change a group pattern,
edit a communication edge, or change information visibility. Deterministic
code applies the operations to a copy of the topology and validates all
roles, references, memberships, nesting, and agent limits. Invalid
proposals use a conservative repair compiler or are skipped.

\subsection{Two-Horizon Playbook Memory}
\label{sec:method-memory}

MANTA maintains memory at two time scales. A short-term playbook lives within a single run and a long-term playbook accumulates across runs.

The short-term playbook logs each turn of the current run, recording the
topology in use, the process flags reported by the Auditor, the repair it
recommended, and the decision the controller took. The Planner therefore
sees which structures have already been tried and which observable
anomalies they produced rather than only the latest state.

The long-term playbook stores general principles that map task characteristics and process risks to topology choices. The Planner consults it both when drafting the initial topology and when choosing a repair. After every $N$ runs a Skill Reflector rewrites its lessons from summaries of recent execution traces.

This learning loop never sees benchmark feedback. Each run in a reflection
batch receives only a process-derived label. A run is called
\emph{procedurally clean} when the audit contains no process flag and
execution ends in decision-grade consensus. This label means that no
encoded coordination anomaly was observed; it does not mean that the
answer is correct. Ground-truth outcomes are used solely for evaluation
and are never exposed to the Planner or either playbook.

%% file: exps.tex
\paragraph{Models and evaluation benchmarks.}
We evaluate MANTA on five benchmarks spanning three complementary capability categories. For \textbf{information seeking and tool use}, we use BrowseComp \citep{wei2025browsecompsimplechallengingbenchmark}, which evaluates multi-step information seeking and evidence synthesis, and StableToolBench \citep{guo-etal-2024-stabletoolbench}, which evaluates reliable external tool selection and execution. For \textbf{planning and workflow execution}, we use PlanCraft \citep{dagan2025plancraft}, which emphasizes long-horizon planning and dependency-aware action sequencing, and WorkBench \citep{styles2024workbench}, which evaluates realistic multi-step workflows. We additionally include MATH \citep{hendrycks2021measuring} as a \textbf{reasoning} benchmark to test whether coordination improvements generalize to tasks that primarily require structured reasoning. All methods use Gemma 4 as the backbone model \citep{gemma4modelcard}. For each benchmark, we evaluate 30 questions and repeat each experiment over three independent runs. 

\paragraph{Baselines.}

\paragraph{Single-Agent Methods.}
\textbf{Single Agent} directly solves each task using one language model agent. \textbf{CoT} solves each task with step-by-step reasoning using a single language model agent \citep{wei2023chainofthoughtpromptingelicitsreasoning,kojima2023largelanguagemodelszeroshot}. \textbf{Self-Consistency} samples multiple reasoning traces and selects the final answer through consistency-based aggregation \citep{wang2023selfconsistencyimproveschainthought}. \textbf{Self-Refine} iteratively improves the answer by generating feedback and revising the response \citep{madaan2023selfrefineiterativerefinementselffeedback}.

\paragraph{Static Multi-Agent Workflows.}
\textbf{Static Agentic Workflows} use fixed communication structures throughout task solving. We evaluate voting, group-chat debate, fully linked debate, orchestrator without discussion, orchestrator with discussion, and an orchestrator tree structure. Among these workflows, multi-agent debate allows multiple agents to propose, critique, and revise candidate answers before producing a final decision \citep{du2023improvingfactualityreasoninglanguage}.

\paragraph{Automatic Agentic System Design Methods.}
\textbf{AFlow} searches over agentic workflows using predefined operators and workflow-level optimization \citep{zhang2025aflowautomatingagenticworkflow}. \textbf{ADAS} uses an LLM-based meta-agent to iteratively propose improved agentic systems based on previous evaluations \citep{hu2025automateddesignagenticsystems}. \textbf{AgentSquare} searches over modular agentic system designs by composing agents, tools, memory, and workflow components \citep{shang2025agentsquareautomaticllmagent}. \textbf{MASS} jointly optimizes agent prompts and communication topologies within a multi-agent design space \citep{zhou2026multiagent}.



\begin{table*}[!t]
\centering
\small

\setlength{\tabcolsep}{2.5pt}
\renewcommand{\arraystretch}{1.08}
\setlength{\aboverulesep}{0.25ex}
\setlength{\belowrulesep}{0.25ex}

\resizebox{0.98\textwidth}{!}{%
\begin{tabular}{@{}llcccccc@{}}
\toprule
\textbf{Category}
& \textbf{System}
& \multicolumn{2}{c}{\textbf{Information Seeking and Tool Use}}
& \multicolumn{2}{c}{\textbf{Planning and Workflow Execution}}
& \textbf{Reasoning}
& \textbf{Average} \\
\cmidrule(lr){3-4}
\cmidrule(lr){5-6}
\cmidrule(lr){7-7}
&
& \textbf{BrowseComp}
& \textbf{StableToolBench}
& \textbf{PlanCraft}
& \textbf{WorkBench}
& \textbf{MATH}
& \\
\midrule

\multirow{4}{*}{\shortstack[l]{Reasoning\\Models}}
& Single Agent
& $34.4_{\pm 4.2}$
& $74.4_{\pm 7.9}$
& $61.1_{\pm 1.6}$
& $41.1_{\pm 5.7}$
& $85.6_{\pm 6.3}$
& $59.3_{\pm 2.5}$ \\

& CoT
& $26.7_{\pm 5.4}$
& $50.0_{\pm 7.2}$
& $62.2_{\pm 12.6}$
& $35.6_{\pm 4.2}$
& $75.6_{\pm 3.1}$
& $50.0_{\pm 3.3}$ \\

& Self-Consistency
& $37.8_{\pm 1.6}$
& $51.1_{\pm 1.6}$
& $61.1_{\pm 15.7}$
& $15.6_{\pm 1.6}$
& $78.9_{\pm 4.2}$
& $48.9_{\pm 3.3}$ \\

& Self-Refine
& $14.4_{\pm 3.1}$
& $68.9_{\pm 1.6}$
& $62.2_{\pm 15.0}$
& $35.6_{\pm 4.2}$
& $\mathbf{96.7}_{\pm 2.7}$
& $55.6_{\pm 3.2}$ \\

\midrule

\multirow{6}{*}{Static MAS}
& Voting
& $43.3_{\pm 2.7}$
& $85.6_{\pm 1.6}$
& $61.1_{\pm 1.6}$
& $41.1_{\pm 1.6}$
& $92.2_{\pm 1.6}$
& $64.7_{\pm 0.8}$ \\

& Group Chat Debate
& $61.1_{\pm 3.1}$
& $82.2_{\pm 5.7}$
& $72.2_{\pm 3.1}$
& $21.1_{\pm 4.2}$
& $91.1_{\pm 4.2}$
& $65.5_{\pm 1.9}$ \\

& Fully Linked Debate
& $58.9_{\pm 9.6}$
& $81.1_{\pm 5.7}$
& $73.3_{\pm 2.7}$
& $21.1_{\pm 4.2}$
& $91.1_{\pm 1.6}$
& $65.1_{\pm 2.5}$ \\

& Orchestrator w/o Discussion
& $53.3_{\pm 2.7}$
& $82.2_{\pm 1.6}$
& $74.4_{\pm 1.6}$
& $23.3_{\pm 4.7}$
& $94.4_{\pm 3.1}$
& $65.5_{\pm 1.3}$ \\

& Orchestrator w/ Discussion
& $64.4_{\pm 4.2}$
& $80.0_{\pm 0.0}$
& $73.3_{\pm 2.7}$
& $20.0_{\pm 2.7}$
& $93.3_{\pm 0.0}$
& $66.2_{\pm 1.1}$ \\

& Orchestrator Tree Structure
& $54.4_{\pm 5.7}$
& $78.9_{\pm 3.1}$
& $62.2_{\pm 3.1}$
& $16.7_{\pm 2.7}$
& $94.4_{\pm 1.6}$
& $61.3_{\pm 1.6}$ \\

\midrule

\multirow{4}{*}{Adaptive MAS}
& AFlow
& $12.2_{\pm 3.1}$
& $66.7_{\pm 5.4}$
& $21.1_{\pm 4.2}$
& $61.1_{\pm 4.2}$
& $\mathbf{96.7}_{\pm 0.0}$
& $51.6_{\pm 1.7}$ \\

& ADAS
& $48.9_{\pm 1.6}$
& $77.8_{\pm 4.2}$
& $57.8_{\pm 13.4}$
& $\mathbf{66.7}_{\pm 0.0}$
& $90.0_{\pm 0.0}$
& $68.2_{\pm 3.2}$ \\

& AgentSquare
& $32.2_{\pm 1.6}$
& $\mathbf{88.9}_{\pm 5.7}$
& $34.4_{\pm 6.8}$
& $62.2_{\pm 3.1}$
& $\mathbf{96.7}_{\pm 2.7}$
& $62.9_{\pm 1.7}$ \\

& MASS
& $50.0_{\pm 2.7}$
& $50.0_{\pm 5.4}$
& $70.0_{\pm 0.0}$
& $46.7_{\pm 0.0}$
& $95.6_{\pm 1.6}$
& $62.5_{\pm 1.2}$ \\

\midrule

\textbf{Ours}
& \textbf{MANTA}\,\mantaicon
& $\mathbf{76.7}_{\pm 4.7}$
& $82.2_{\pm 3.1}$
& $\mathbf{76.7}_{\pm 3.3}$
& $43.3_{\pm 2.7}$
& $91.1_{\pm 5.7}$
& $\mathbf{74.0}_{\pm 1.8}$ \\

\bottomrule
\end{tabular}%
}

\caption{
Main results across five benchmarks using Gemma 4 31B with medium
reasoning effort. Methods are grouped into reasoning models, static MAS,
and adaptive MAS. We report mean success rates over three runs, with
standard deviations in subscript. Best results, including ties, are bolded.
}
\label{tab:main_results}
\end{table*}

%% file: discussion.tex
\subsection{Experimental Results and Analysis}



\paragraph{Overall performance.}
Table~\ref{tab:main_results} shows MANTA performs consistently across
information seeking, tool use, planning, workflow execution, and reasoning.
\textbf{MANTA achieves the strongest average score of $74.0$ across the five
benchmarks, leading the next best method by $5.8$ points.} It also improves
on the Single Agent across every benchmark. The results indicate 
adapting the collaboration to each task provides a broad advantage rather
than a gain limited to one task type.

The supporting studies explain where this advantage comes from. The ablation
study identifies strong initial planning as the largest contributor, while
repair and past experience provide additional improvements. The mutation
analysis shows the first change captures most of the benefit, with up to
three changes extending coverage to the most difficult tasks. Past learned experience also improves results on the same benchmark and produces the only positive average gain when transferred across benchmarks. At the same time, MANTA uses the fewest tokens among the evaluated multi-agent systems. \textbf{Together,
these findings show that careful planning, targeted changes, and reusable experience improve overall performance without requiring more computation.}

\tikzset{
repairagent/.style={
circle,
draw=black!65,
fill=white,
minimum size=4.8mm,
inner sep=0pt,
font=\tiny,
line width=0.6pt
},
repaircoord/.style={
repairagent,
fill=cplan!25,
draw=cplan!75
},
repairverif/.style={
repairagent,
fill=caudit!30,
draw=caudit!80
},
repairissue/.style={
draw=caudit!80,
fill=caudit!8,
rounded corners=3pt,
align=center,
font=\scriptsize,
inner sep=4pt,
text width=2.15cm
},
repairaction/.style={
draw=cmut!80,
fill=cmut!7,
rounded corners=3pt,
align=center,
font=\scriptsize,
inner sep=4pt,
text width=2.15cm
},
repairtopo/.style={
draw=black!35,
dashed,
rounded corners=3pt,
inner sep=3pt
},
repairflow/.style={
-{Stealth[length=2mm]},
draw=black!65,
line width=0.75pt
},
mutationflow/.style={
-{Stealth[length=2mm]},
draw=cmut!85!black,
line width=0.85pt
},
repairlabel/.style={
font=\tiny,
align=center,
text=black!75
}
}
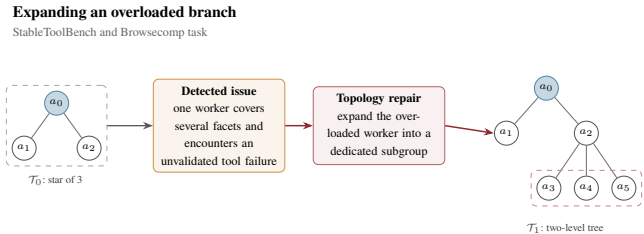
\begin{figure}[!htbp]
\centering

\resizebox{\columnwidth}{!}{%
\begin{tikzpicture}

\node[
font=\small\bfseries,
anchor=west
] at (0,5.05)
{Expanding an overloaded branch};

\node[
font=\scriptsize,
anchor=west,
text=black!60
] at (0,4.65)
{StableToolBench and Browsecomp task};

\node[repaircoord] (a0) at (1.0,3.25) {$a_0$};
\node[repairagent] (a1) at (0.35,2.35) {$a_1$};
\node[repairagent] (a2) at (1.65,2.35) {$a_2$};

\draw[black!55, line width=0.6pt]
(a0)--(a1)
(a0)--(a2);

\node[
repairtopo,
fit=(a0)(a1)(a2)
] (initial) {};

\node[
repairlabel,
below=2pt of initial
] {
$\mathcal{T}_0$: star of 3
};

\node[
repairissue,
text width=2.35cm
] (issue) at (4.25,2.8) {
\textbf{Detected issue}\\
one worker covers several facets and encounters
an unvalidated tool failure
};

\node[
repairaction,
text width=2.35cm
] (repair) at (7.45,2.8) {
\textbf{Topology repair}\\
expand the overloaded worker into a dedicated subgroup
};

\node[repaircoord] (b0) at (10.8,3.55) {$a_0$};
\node[repairagent] (b1) at (10.0,2.65) {$a_1$};
\node[repairagent] (b2) at (11.6,2.65) {$a_2$};

\node[repairagent] (b3) at (10.85,1.55) {$a_3$};
\node[repairagent] (b4) at (11.6,1.55) {$a_4$};
\node[repairagent] (b5) at (12.35,1.55) {$a_5$};

\draw[black!55, line width=0.6pt]
(b0)--(b1)
(b0)--(b2)
(b2)--(b3)
(b2)--(b4)
(b2)--(b5);

\node[
repairtopo,
draw=cmut!60,
fit=(b3)(b4)(b5)
] {};

\node[
repairlabel
] at (11.2,0.75) {
$\mathcal{T}_1$: two-level tree
};

\draw[repairflow]
(initial.east) --
(issue.west);

\draw[mutationflow]
(issue.east) --
(repair.west);

\draw[mutationflow]
(repair.east) --
(b1.west);

\end{tikzpicture}%
}

\caption{\textbf{Branch expansion.}
When one worker in the initial star becomes overloaded and encounters an unvalidated tool failure, MANTA expands that worker into a dedicated subgroup while preserving the unaffected branch.}
\label{fig:manta_branch_expansion}
\end{figure}



\subsection{Ablation Study}
\label{sec:ablation}

We evaluate four variant versions of MANTA to measure the contribution
of its main components. This experiment uses 30 tasks from each of
BrowseComp, WorkBench, PlanCraft, and StableToolBench. Table~\ref{tab:manta_ablation} reports the average success
rate across these four benchmarks.


\begin{table}[t]
\centering
\scriptsize
\setlength{\tabcolsep}{3pt}
\renewcommand{\arraystretch}{1.05}
\resizebox{\columnwidth}{!}{%
\begin{tabular}{@{}lrrrr@{}}
\toprule
& Success & \multicolumn{3}{c}{Mean tokens per run} \\
\cmidrule(lr){3-5}
Ablation setting & (\%) & Input & Output & Total \\
\midrule
\textbf{Full MANTA} & \textbf{71.7} & 94{,}811 & 5{,}504 & 100{,}315 \\
No initial Topology Planner & 57.5 & 105{,}941 & 6{,}099 & 112{,}040 \\
No topology mutation & 60.8 & \textbf{67{,}528} & 4{,}577 & \textbf{72{,}105} \\
No long-term playbook update & 67.5 & 97{,}145 & 5{,}474 & 102{,}620 \\
No long-term playbook & 66.7 & 74{,}067 & \textbf{4{,}289} & 78{,}356 \\
\bottomrule
\end{tabular}
}
\caption{MANTA ablation results averaged across four benchmarks. Token
counts include the amortized cost of batched long-term playbook updates.}
\label{tab:manta_ablation}
\end{table}

The first ablation replaces task-conditioned initial topology planning
with a fixed coordinator--worker topology; it retains the Auditor and the
repair-time mutation planner. The second retains the initial Topology
Planner, but disables audit-triggered topology mutation. The remaining two settings either freeze updates to the
long-term playbook or remove the long-term playbook entirely.

Full MANTA achieves the highest average success rate. Replacing
task-conditioned planning with a fixed topology causes the largest drop,
from 71.7 to 57.5. Removing topology repair reduces the average to 60.8.
Freezing or removing the long-term playbook gives smaller but consistent
drops. These results show that initial topology planning and execution-time
repair provide the largest gains, while cross-run learning provides an
additional benefit.

\subsection{Effect of Mutation Budget}
\label{sec:mutation_budget}


We next test whether a larger topology-mutation budget helps MANTA solve
more tasks. We run full MANTA with budgets from zero to three on 30 tasks
from each of BrowseComp, StableToolBench, PlanCraft, and WorkBench. We evaluate the budgets progressively.
Starting at zero, we retain solved tasks and isolate the unsuccessful ones.
At each larger budget, we identify failures that flip to success and carry
those successes forward. Thus, at budget $b$, a task is counted as solved
if it has succeeded at any evaluated budget up to $b$. This measures the
task coverage provided by the available repair budget.


\begin{figure}[t]
\centering
\includegraphics[width=\columnwidth]{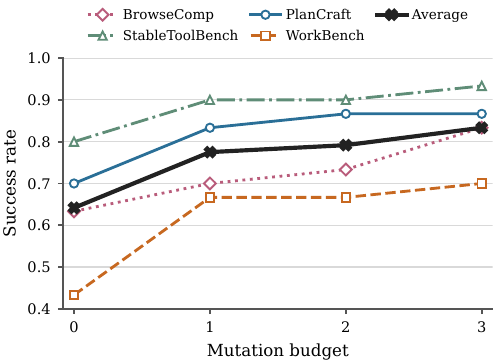}
\caption{Success rate under different mutation budgets.}
\label{fig:mutation_budget}
\end{figure}

As shown in Figure~\ref{fig:mutation_budget}, all four benchmarks improve as the
available budget increases. The largest average gain comes from the first
mutation opportunity, while later mutations provide smaller additional
gains. These results show that bounded topology repair expands the set of
tasks MANTA can solve.

\subsection{Transferability of the Long-Term Playbook}
\label{sec:playbook_transfer}

A central goal of MANTA's long-term playbook is to accumulate structural
experience that remains useful beyond the tasks on which it was originally
acquired. We therefore evaluate whether the learned playbook can improve
topology initialization on unseen tasks from either the same benchmark or a
different benchmark. This experiment examines whether MANTA learns reusable
structural knowledge rather than benchmark-specific workflows.

We construct a playbook from 30 source-benchmark runs and update it after
every 10 runs. The resulting playbook is then frozen and evaluated on 30
target tasks. In-domain transfer uses the same benchmark for learning and
evaluation, whereas cross-domain transfer exchanges PlanCraft and WorkBench
as the source and target. We set the mutation budget to zero so that any
performance difference comes solely from playbook-guided topology
initialization rather than execution-time topology adaptation.

\begin{table}[t]
\centering
\scriptsize
\setlength{\tabcolsep}{3pt}
\renewcommand{\arraystretch}{1.08}
\resizebox{\columnwidth}{!}{%
\begin{tabular}{@{}lccc@{}}
\toprule
\multicolumn{4}{c}{\textbf{Cross-domain transfer}} \\
\midrule
\textbf{Method}
& \textbf{PlanCraft $\rightarrow$ WorkBench}
& \textbf{WorkBench $\rightarrow$ PlanCraft}
& \textbf{Mean $\Delta$} \\
\midrule

ADAS
& $66.7 \rightarrow 66.7$
& $76.7 \rightarrow 70.0$
& $-3.3$ \\

AgentSquare
& $60.0 \rightarrow 36.7$
& $26.7 \rightarrow 23.3$
& $-13.3$ \\

MASS
& $46.7 \rightarrow 0.0$
& $70.0 \rightarrow 0.0$
& $-58.3$ \\

\textbf{MANTA}
& $\mathbf{43.3 \rightarrow 50.0}$
& $\mathbf{70.0 \rightarrow 70.0}$
& $\mathbf{+3.3}$ \\
\midrule

\multicolumn{4}{c}{\textbf{In-domain playbook adaptation (MANTA)}} \\
\midrule
\textbf{Method}
& \textbf{PlanCraft $\rightarrow$ PlanCraft}
& \textbf{WorkBench $\rightarrow$ WorkBench}
& \textbf{Mean $\Delta$} \\
\midrule

\textbf{MANTA} \mantaicon
& $\mathbf{70.0 \rightarrow 73.3}$
& $\mathbf{43.3 \rightarrow 46.7}$
& $\mathbf{+3.3}$ \\
\bottomrule
\end{tabular}%
}
\caption{Playbook transfer at mutation budget zero. Each cell reports the success rate before and after transferring source-benchmark experience. The pre-transfer success rate is the median across three runs (\%).}
\label{tab:playbook_transfer}
\end{table}

As shown in Table~\ref{tab:playbook_transfer}, MANTA benefits from both
in-domain and cross-domain experience. In-domain reuse improves PlanCraft
and WorkBench by $3.3$ points each, while cross-domain transfer yields a
mean gain of $3.3$ points. In contrast, transferring the structures learned
by the adaptive baselines produces neutral or substantially negative effects.
MANTA transfers more effectively because its long-term playbook preserves
inheritable and actionable structural knowledge rather than fixed workflows
optimized for predefined training tasks. Since MANTA does not assume a
specific training-task set, these results demonstrate its ability to
dynamically adapt its topology to different tasks and domains.

\subsection{Token Usage}
\label{app:token_usage}

Although token efficiency is not the primary objective of MANTA, the
results reveal an additional benefit of adaptive topology revision.
As shown in Table~\ref{tab:token_usage}, static multi-agent systems often
incur substantial inference costs because they execute fixed agent
configurations and communication patterns regardless of task needs.
Meanwhile, workflow-optimization methods can reduce inference-time usage,
but introduce additional offline search or validation costs. MANTA
achieves the lowest overall token consumption among the evaluated
multi-agent systems. In particular, its meta-level operations account for
only approximately $12\%$ of its inference budget, suggesting that
targeted topology planning and repair can improve coordination without
introducing substantial token overhead.

\subsection{Topology Evolution and Repair}
\label{sec:results-repair} 
\begin{table}[t]
\centering
\scriptsize
\setlength{\tabcolsep}{2.2pt}
\renewcommand{\arraystretch}{0.90}
\setlength{\aboverulesep}{0.20ex}
\setlength{\belowrulesep}{0.20ex}

\resizebox{\columnwidth}{!}{%
\begin{tabular}{@{}llrrrrrrr@{}}
\toprule
\textbf{Type}
& \textbf{System}
& \multicolumn{3}{c}{\textbf{Offline}}
& \multicolumn{3}{c}{\textbf{Inference}}
& \textbf{Overall} \\
\cmidrule(lr){3-5}
\cmidrule(lr){6-8}
&
& \textbf{In}
& \textbf{Out}
& \textbf{Total}
& \textbf{In}
& \textbf{Out}
& \textbf{Total}
& \textbf{Total} \\
\midrule

Single
& Single Agent
& \multicolumn{3}{c}{\textemdash}
& 17{,}767
& 4{,}044
& 21{,}811
& 21{,}811 \\

\midrule

\multirow{6}{*}{Static}
& Voting
& \multicolumn{3}{c}{\multirow{6}{*}{\textemdash}}
& 64{,}895
& 15{,}885
& 80{,}781
& 80{,}781 \\

& Orch. (No Disc.)
& \multicolumn{3}{c}{}
& 94{,}125
& 21{,}173
& 115{,}298
& 115{,}298 \\

& Orch. (Tree)
& \multicolumn{3}{c}{}
& 107{,}227
& 28{,}194
& 135{,}421
& 135{,}421 \\

& Orch. (Discussion)
& \multicolumn{3}{c}{}
& 149{,}950
& 34{,}257
& 184{,}207
& 184{,}207 \\

& Fully Linked
& \multicolumn{3}{c}{}
& 128{,}726
& 34{,}212
& 162{,}938
& 162{,}938 \\

& Group Chat
& \multicolumn{3}{c}{}
& 159{,}809
& 41{,}843
& 201{,}651
& 201{,}651 \\

\midrule

\multirow{4}{*}{Adaptive}
& AFlow
& 126{,}292
& 6{,}225
& 132{,}517
& 17{,}472
& 1{,}082
& 18{,}553
& 151{,}070 \\

& ADAS
& 229{,}228
& 18{,}447
& 247{,}675
& 26{,}220
& 1{,}508
& 27{,}728
& 275{,}403 \\

& AgentSquare
& 115{,}561
& 27{,}494
& 143{,}056
& 13{,}009
& 1{,}929
& 14{,}938
& 157{,}993 \\

& MASS
& 49{,}801
& 3{,}347
& 53{,}148
& 120{,}690
& 7{,}188
& 127{,}878
& 181{,}026 \\

\midrule

\textbf{Ours}
& \textbf{MANTA}\,\mantaicon
& \multicolumn{3}{c}{\textemdash}
& \textbf{63{,}724}
& \textbf{13{,}928}
& \textbf{77{,}652}
& \textbf{77{,}652} \\

\bottomrule
\end{tabular}%
}

\caption{\textbf{Mean token usage per run across four benchmarks.}
Offline costs correspond to workflow search or validation;
MANTA's inference usage includes 9{,}416 meta-level and
68{,}236 inner-agent tokens.}
\label{tab:token_usage}
\end{table}

\paragraph{Expanding overloaded branches.}
Figure~\ref{fig:manta_branch_expansion} shows a StableToolBench and Browsecomp task with multiple retrieval facets. The Planner initially selects a three-agent star in which a coordinator distributes the facets across two workers. During execution, one worker becomes responsible for several facets and encounters a tool failure without sufficient downstream validation. MANTA responds by expanding that worker into the hub of a dedicated subgroup. The topology evolves from a star into a two-level tree, increasing decomposition and specialization only within the affected branch.

\begin{figure}[!htbp]
\centering

\resizebox{\columnwidth}{!}{%
\begin{tikzpicture}

\node[
font=\small\bfseries,
anchor=west
] at (0,5.2)
{Repairing verification and execution structure};

\node[
font=\scriptsize\bfseries,
anchor=west,
text=black!75
] at (0,4.65)
{Missing validation};

\node[repairagent] (c0) at (1.0,3.65) {$a_0$};

\node[
repairlabel
] at (1.0,3.05)
{singleton};

\node[
repairissue
] (issue1) at (4.15,3.65) {
low confidence\\
no downstream check
};

\node[
repairagent
] (d0) at (7.15,3.65) {$a_0$};

\node[
repairverif
] (d1) at (8.25,3.65) {$a_1$};

\draw[repairflow]
(d0) -- (d1);

\node[
repairlabel
] at (7.7,3.05)
{proposer $\rightarrow$ verifier};

\draw[repairflow]
(c0) -- (issue1.west);

\draw[mutationflow]
(issue1.east) -- (d0.west);

\draw[
black!20,
dashed,
line width=0.6pt
] (0,2.35) -- (9.2,2.35);

\node[
font=\scriptsize\bfseries,
anchor=west,
text=black!75
] at (0,1.95)
{Duplicate state-changing actions};

\node[repaircoord] (e0) at (1.0,0.95) {$a_0$};
\node[repairagent] (e1) at (0.4,0.15) {$a_1$};
\node[repairverif] (e2) at (1.6,0.15) {$a_2$};

\draw[black!55, line width=0.6pt]
(e0)--(e1)
(e0)--(e2);

\node[
repairlabel
] at (1.0,-0.45)
{parallel star};

\node[
repairissue
] (issue2) at (4.15,0.55) {
repeated write\\
from parallel agents
};

\node[repairagent] (f0) at (6.75,0.55) {$a_0$};
\node[repairagent] (f1) at (7.8,0.55) {$a_1$};
\node[repairverif] (f2) at (8.85,0.55) {$a_2$};

\draw[repairflow]
(f0) -- (f1);

\draw[repairflow]
(f1) -- (f2);

\node[
repairlabel
] at (7.8,-0.05)
{serialized chain};

\draw[repairflow]
(e2.east) -- (issue2.west);

\draw[mutationflow]
(issue2.east) -- (f0.west);

\end{tikzpicture}%
}

\caption{\textbf{Verification and serialization repairs.}
Missing validation introduces a verifier, while duplicated state-changing actions replace parallel execution with a serialized chain.}
\label{fig:manta_structure_repairs}
\end{figure}
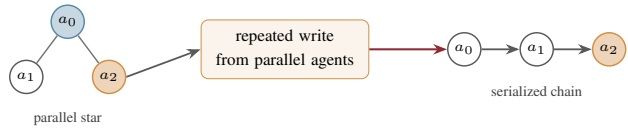

\paragraph{Changing verification and execution order.}
Figure~\ref{fig:manta_structure_repairs} shows two repairs that modify coordination without increasing the size of the entire system. When a singleton produces a low-confidence output without verification, MANTA adds a verifier and forms a two-agent chain. When parallel agents attempt the same state-changing action, MANTA replaces the star with a chain that serializes execution and prevents the action from being applied more than once.

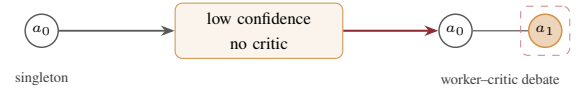
\begin{figure}[!htbp]
\centering

\resizebox{\columnwidth}{!}{%
\begin{tikzpicture}

\node[
font=\small\bfseries,
anchor=west
] at (0,5.2)
{Rewiring communication and inserting a critic};

\node[
font=\scriptsize\bfseries,
anchor=west,
text=black!75
] at (0,4.65)
{Premature consensus};

\node[repaircoord] (s0) at (1.0,3.85) {$a_0$};
\node[repairagent] (s1) at (0.4,3.05) {$a_1$};
\node[repairverif] (s2) at (1.6,3.05) {$a_2$};

\draw[black!55, line width=0.6pt]
(s0)--(s1)
(s0)--(s2);

\node[
repairlabel
] at (1.0,2.45)
{star of 3};

\node[
repairissue
] (issue1) at (4.15,3.45) {
agents agree early\\
open issues remain
};

\node[repaircoord] (d0) at (7.4,3.85) {$a_0$};
\node[repairagent] (d1) at (6.8,3.05) {$a_1$};
\node[repairverif] (d2) at (8.0,3.05) {$a_2$};

\draw[black!55, line width=0.6pt]
(d0)--(d1)
(d0)--(d2);

\draw[cmut!70, line width=0.9pt]
(d1)--(d2);

\node[
repairlabel
] at (7.4,2.45)
{debate of 3};

\draw[repairflow]
(s2.east) -- (issue1.west);

\draw[mutationflow]
(issue1.east) -- (d1.west);

\draw[
black!20,
dashed,
line width=0.6pt
] (0,2.1) -- (9.2,2.1);

\node[
font=\scriptsize\bfseries,
anchor=west,
text=black!75
] at (0,1.7)
{Unchecked answer};

\node[repairagent] (w0) at (1.0,0.75) {$a_0$};

\node[
repairlabel
] at (1.0,0.05)
{singleton};

\node[
repairissue
] (issue2) at (4.15,0.75) {
low confidence\\
no critic
};

\node[repairagent] (g0) at (7.0,0.75) {$a_0$};
\node[repairverif] (g1) at (8.3,0.75) {$a_1$};

\draw[black!55, line width=0.6pt]
(g0)--(g1);

\node[
repairtopo,
draw=cmut!60,
fit=(g1)
] {};

\node[
repairlabel
] at (7.65,0.05)
{worker--critic debate};

\draw[repairflow]
(w0.east) -- (issue2.west);

\draw[mutationflow]
(issue2.east) -- (g0.west);

\end{tikzpicture}%
}

\caption{\textbf{Rewiring and critic insertion.}
Illustration of two topology repairs that add a missing communication edge to transform a star into a fully connected debate and insert a critic to transform a singleton into a worker--critic debate. Highlighted elements indicate the modifications introduced by each repair.}
\label{fig:manta_trace_mutations}

\end{figure}

\paragraph{Rewiring communication and inserting a critic.}
The traces contain two further repairs, and neither one grows a branch
(Figure~\ref{fig:manta_trace_mutations}). The first rewires communication
while keeping the agent set fixed. In the PlanCraft trace, all three agents agreed after the first turn, but the
agreement came with imperfect confidence and issues still open, so the auditor
flagged it as premature consensus. The repair kept all three agents and added
the one edge the star lacked, a direct link between worker and verifier,
turning the topology into a fully connected debate
(Figure~\ref{fig:manta_trace_mutations}, top). The next turn passed the audit,
terminated by consensus with confidence $1.0$, and the answer received
benchmark score of $1.0$.

The second repair adds validation capacity instead of another solver. In the
Math500 trace, a single agent returned a
low-confidence answer that no one checked. The repair inserted exactly one
agent, assigned it the critic role, and paired the two in a debate
(Figure~\ref{fig:manta_trace_mutations}, bottom). The next audit was again
clean, and the run ended with a consensus confidence of $1.0$ and a score of $1.0$. The
added agent never attempts the task itself; its only job is to check the
answer that already exists.

\medskip
\noindent
\fcolorbox{black!55}{black!3}{%
\parbox{\dimexpr\columnwidth-2\fboxsep-2\fboxrule\relax}{%
\small
\textbf{Insight.}
Topology evolution is not simply adding agents or edges. Of the five
repairs above, only branch expansion makes the system larger:
serialization reorders execution, rewiring changes who talks to whom, and
critic insertion gives new compute a checking role rather than a solving
one. In every case, the appropriate repair is determined by the structural
failure visible in the execution trace, not by the size of the system.
}%
}
\medskip

\subsection{How Well Does Trace Auditing Work?}
\label{sec:results-audit}

We evaluate the initial audit over all $450$ MANTA runs without exposing
benchmark verdicts to the system. Runs with no repair flag are
correct $83.2$ percent of the time, compared with $62.5$ percent for
flagged runs. The resulting $20.7$ point separation shows that trace
quality provides meaningful information about answer reliability while
remaining independent of the benchmark answer. The Auditor identifies
$75$ of the $117$ incorrect runs. On WorkBench it reaches F1 $0.78$
because empty branches and duplicated state changes leave clear process
evidence. It rarely changes PlanCraft and MATH runs, which preserves the
original reasoning process when no structural symptom is visible.

MANTA applies $151$ repairs. The number of process flags decreases after
$60.9$ percent of repairs, and $59.6$ percent of named repair targets are
absent from the next audit. Repaired runs produce a correct final answer
in $69.5$ percent of cases. A clean trace remains a reliability signal
rather than a correctness guarantee, since agents can agree on the same
incorrect answer without an observable process anomaly. 

%% file: appendix.tex
\newtcolorbox{promptbox}[1]{%
  breakable, enhanced,
  colback=black!4, colframe=black!60, boxrule=0.5pt, arc=1.5pt,
  left=6pt, right=6pt, top=5pt, bottom=5pt,
  title={#1},
  fonttitle=\footnotesize\bfseries\sffamily,
  colbacktitle=black!12, coltitle=black,
  fontupper=\footnotesize\sffamily,
  before upper={\setlength{\parindent}{0pt}\setlength{\parskip}{3.5pt}},
}
\newcommand{\var}[1]{{\normalfont\itshape$\langle$#1$\rangle$}}
\newcommand{\code}[1]{\textsf{#1}}

\appendix
\section{Implementation Details}
\label{app:implementation_details}

\subsection{Experimental Setup and Fair Comparison}
For fair comparison, all multi-agent methods are constrained to the same
maximum number of agents and comparable inference budget. Static workflows
use the same set of agent roles as MANTA but keep their communication
topology fixed throughout the entire task-solving process. For
optimization-based baselines such as AFlow, ADAS, AgentSquare, and MASS, we
follow their original optimization procedures when applicable and match the
search budget as closely as possible. For MANTA, topology revision is
performed online according to intermediate collaboration signals, and the
revised topology is used for subsequent reasoning steps. Unless otherwise
stated, the final answer is produced by the designated aggregator or
coordinator agent.

\subsection{MANTA Configuration}
\label{app:manta_config}
Table~\ref{tab:manta_hparams} lists the configuration used in all MANTA
experiments. All meta-agents (Topology Planner, Trace Auditor, and Skill
Reflector) use the same backbone model as the task agents, with temperature
$0$. If the Planner returns an invalid or unparseable plan, the
Orchestrator substitutes a deterministic fallback topology (a
coordinator--worker star, or a singleton when the budget is one agent).

\medskip
\noindent\begin{minipage}{\columnwidth}
\centering
\small
\begin{tabular}{@{}p{0.56\columnwidth}p{0.34\columnwidth}@{}}
\toprule
\textbf{Parameter} & \textbf{Value} \\
\midrule
Maximum initial agents & 5 \\
Maximum total agents (after repair) & 10 \\
Topology mutations per run & 1 \\
Operations per mutation & $\leq 3$ \\
Root interaction patterns & singleton, star, chain, debate, voting \\
Meta-agent temperature & 0.0 \\
long-term playbook reflection batch size & 12 runs \\
\bottomrule
\end{tabular}
\captionof{table}{MANTA hyperparameters.}
\label{tab:manta_hparams}
\end{minipage}
\medskip

\subsection{Topology Mutation Operators}
\label{app:mutation_ops}
For a non-retrieval task, a repair is expressed as at most three of the
operators in Table~\ref{tab:mutation_ops}. The Planner emits a compact JSON
object containing a rationale and an ordered operation list. Deterministic
code discards unknown operators, normalizes role aliases, and attaches the
current audit modes as the mutation targets. It then applies the operations
to a copy of the current topology specification. The new version is
accepted only if it passes full structural validation (unique group
membership, valid identifiers and roles, acyclic attachment structure,
legal group patterns, and the maximum agent budget); otherwise the proposal
is rejected before any agent is instantiated. A conservative repair
compiler may then map the grounded audit modes to one local edit; if it
cannot do so, execution finalizes without a mutation.

\medskip
\noindent\begin{minipage}{\columnwidth}
\centering
\small
\begin{tabular}{@{}p{0.34\columnwidth}p{0.56\columnwidth}@{}}
\toprule
\textbf{Operator} & \textbf{Effect} \\
\midrule
\code{add\_agent} & Add one agent to an existing group, with a specified
structural role (e.g., verifier) and stage role (worker or critic). \\
\addlinespace[2pt]
\code{expand\_agent\_\allowbreak to\_group} & Turn one agent into the hub
of a new nested subgroup with its own interaction pattern and members. \\
\addlinespace[2pt]
\code{set\_group\_\allowbreak pattern} & Change the interaction pattern of
an existing group (e.g., star $\rightarrow$ debate, star $\rightarrow$
chain). \\
\addlinespace[2pt]
\code{add\_edge},\newline\code{remove\_edge} & Add or remove a direct
communication link between two agents. \\
\addlinespace[2pt]
\code{set\_context\_\allowbreak policy} & Adjust an agent's information
visibility (e.g., grant global access to the shared evidence ledger). \\
\bottomrule
\end{tabular}
\captionof{table}{Bounded topology mutation operators available to the
Planner during trace-backed repair.}
\label{tab:mutation_ops}
\end{minipage}
\medskip

Retrieval-heavy tasks use the same audit and budget gate but a separate
deterministic mutation. The resource guard selects the existing
non-coordinator that made the most search calls (with deterministic
tie-breaking), replaces the active topology by a singleton containing that
agent, grants it global evidence access, and assigns a focused
evidence-recovery directive. This \emph{retrieval contraction} does not add
agents and is not part of the Planner's mutation language. It avoids
re-running a wide set of searchers after the first fan-out has already
failed. All old packets, evidence, and turn-level candidates remain in the
append-only run state under both repair paths.

\subsection{Deterministic Audit Taxonomy}
\label{app:audit_taxonomy}
The deterministic component of the Trace Auditor scans the execution trace
for the fixed process-anomaly patterns in
Table~\ref{tab:audit_taxonomy}. The implementation calls these patterns
``failure modes,'' but the term is diagnostic: a match means that an
encoded risk is visible in the trace, not that the final answer is
incorrect. Each flag carries a severity, the implicated agents, and a
repairability field; the open-set LLM auditor may add further
trace-grounded flags beyond this taxonomy.

\medskip
\noindent\begin{minipage}{\columnwidth}
\centering
\small
\begin{tabular}{@{}p{0.42\columnwidth}p{0.48\columnwidth}@{}}
\toprule
\textbf{Process flag} & \textbf{Signal in the trace} \\
\midrule
\code{tool\_error\_\allowbreak cascade} & Repeated failing tool calls
within a branch. \\
\addlinespace[2pt]
\code{branch\_collapse} & A branch yields no substantive artifact
(blocked or empty output). \\
\addlinespace[2pt]
\code{unsupported\_\allowbreak impossibility\_claim} & An agent declares
the task unanswerable without supporting evidence. \\
\addlinespace[2pt]
\code{unverified\_\allowbreak impossibility\_\allowbreak consensus} &
Agents agree the task is unanswerable without independent verification. \\
\addlinespace[2pt]
\code{evidence\_lost\_\allowbreak before\_synthesis} & Gathered evidence
is absent from the synthesis input. \\
\addlinespace[2pt]
\code{premature\_\allowbreak consensus} & Agreement despite low confidence
or open unresolved issues. \\
\addlinespace[2pt]
\code{message\_\allowbreak compaction\_loss} & Relay-packet compaction
dropped load-bearing content. \\
\addlinespace[2pt]
\code{insufficient\_\allowbreak search\_coverage} & Too few distinct
queries for a broad retrieval task. \\
\addlinespace[2pt]
\code{duplicate\_state\_\allowbreak mutation} & The same state-changing
tool call is issued by multiple agents. \\
\addlinespace[2pt]
\code{missing\_validator} & A high-precision task lacks a distinct
verification step. \\
\addlinespace[2pt]
\code{give\_up\_shaped\_\allowbreak candidate} & Every candidate answer
concludes that the task cannot be completed. \\
\bottomrule
\end{tabular}
\captionof{table}{Deterministic process-anomaly taxonomy used by the
Trace Auditor. Internal identifiers retain the implementation's
\code{failure\_mode} vocabulary.}
\label{tab:audit_taxonomy}
\end{minipage}
\medskip

\subsection{Trace Auditing Statistics}
\label{app:audit_stats}
This section is the aggregate results of Trace Auditing over the $450$ MANTA runs.

\paragraph{Correctness-proxy scores per benchmark.}
Table~\ref{tab:audit_perbench} compares the \emph{initial} repair-level
audit flag with benchmark answer incorrectness. Using the first audit
avoids evaluating a detector after its own intervention. Here precision,
recall, FPR, and FNR use answer incorrectness as the positive class. They
must not be read as literal process-detection scores: no human label says
whether a trace truly contains a coordination defect. ``False positive''
therefore means only \emph{flagged but ultimately answer-correct}, and
``false negative'' means \emph{clean but answer-incorrect}.

Across all runs, the audit flags $64.1$ percent of incorrect answers at
$37.5$ percent precision. The aggregate FPR and FNR are $37.5$ and $35.9$
percent. The benchmark spread is more informative than the aggregate:
WorkBench answer errors often co-occur with visible workflow anomalies,
whereas PlanCraft and MATH frequently contain wrong reasoning with no
observable coordination symptom. BrowseComp and StableToolBench show the
opposite difficulty: retrieval or tool-use anomalies are common but often
recoverable, producing high proxy FPR.

\medskip
\noindent\begin{minipage}{\columnwidth}
\centering
\small
\setlength{\tabcolsep}{3.2pt}
\begin{tabular}{@{}lrrrrrr@{}}
\toprule
\textbf{Benchmark} & \textbf{Flag} & \textbf{Prec.} & \textbf{Rec.} &
\textbf{FPR} & \textbf{FNR} & \textbf{F1} \\
\midrule
BrowseComp & 83 & 0.25 & 1.00 & 0.90 & 0.00 & 0.40 \\
StableToolBench & 67 & 0.21 & 0.88 & 0.72 & 0.13 & 0.34 \\
PlanCraft & \phantom{0}1 & 1.00 & 0.05 & 0.00 & 0.95 & 0.09 \\
WorkBench & 47 & 0.81 & 0.75 & 0.23 & 0.25 & 0.78 \\
MATH & \phantom{0}2 & 0.50 & 0.13 & 0.01 & 0.88 & 0.20 \\
\midrule
All & 200 & 0.38 & 0.64 & 0.38 & 0.36 & 0.47 \\
\bottomrule
\end{tabular}
\captionof{table}{Initial audit as a proxy for answer incorrectness, with
$90$ runs per benchmark. Flag is the number of runs whose audit
recommends repair. Metrics treat benchmark-incorrect as positive; they do
not use manually annotated process-failure ground truth.}
\label{tab:audit_perbench}
\end{minipage}
\medskip

\noindent\begin{minipage}{\columnwidth}
\centering
\small
\begin{tabular}{@{}lrr@{}}
\toprule
\textbf{Initial audit} & \textbf{Incorrect} & \textbf{Correct} \\
\midrule
Repair flag & 75 & 125 \\
No repair flag & 42 & 208 \\
\midrule
Total & 117 & 333 \\
\bottomrule
\end{tabular}
\captionof{table}{Initial audit outcome versus benchmark correctness over
all $450$ MANTA runs.}
\label{tab:audit_detection}
\end{minipage}
\medskip

\paragraph{Which process flags correlate with correctness.}
Table~\ref{tab:audit_modes} reports, for each mode, how many runs it
appears in, the difference in success rate between runs where it fires
and runs where it does not, and how often a repair that targeted it
removed it from the following audit. Seven of the nine modes are
associated with a lower success rate, and the four strongest of them
carry a gap of more than $30$ points, showing that several deterministic
detectors contain useful correctness information. This association does
not establish that every firing is a true process defect. Two
modes, relay-packet compaction loss and failing tool calls, show no such
gap. They capture recoverable events that a competent topology absorbs,
and down-weighting them in the severity gate is a direct way to raise
precision in future work.

\medskip
\noindent\begin{minipage}{\columnwidth}
\centering
\small
\begin{tabular}{@{}p{0.40\columnwidth}ccc@{}}
\toprule
\textbf{Process flag} & \textbf{Runs} & \textbf{$\Delta$ succ.} &
\textbf{Cleared} \\
\midrule
Evidence lost before synthesis & 7 & $-60.7$ & 6/6 \\
Missing validator & 71 & $-46.1$ & 17/19 \\
Branch collapse & 100 & $-38.6$ & 4/12 \\
Give-up shaped candidate & 36 & $-32.1$ & n/a \\
Premature consensus & 48 & $-17.5$ & 9/11 \\
Duplicate state mutation & 143 & $-14.2$ & 23/39 \\
Insufficient search coverage & 68 & $-5.8$ & n/a \\
Message compaction loss & 131 & $+2.2$ & 5/8 \\
Tool error cascade & 61 & $+5.4$ & 23/51 \\
\bottomrule
\end{tabular}
\captionof{table}{Detected process flags over all $450$ runs. $\Delta$
succ. is the success rate of runs containing the mode minus the success
rate of runs without it, in percentage points, against an overall rate of
$74.0$. Cleared is the share of repairs targeting that mode after which
the mode is absent from the next audit. Modes marked n/a are only
observed on retrieval runs, which are repaired by a deterministic
contraction that names no target.}
\label{tab:audit_modes}
\end{minipage}
\medskip

\paragraph{Effect of repair in detail.}
The pre/post comparison is made on audit outputs, not benchmark
counterfactuals. Of $151$ repairs, $92$ ($60.9\%$) reduce the total number
of flags, $27$ ($17.9\%$) leave it unchanged, and $32$ ($21.2\%$)
increase it; $27$ ($17.9\%$) leave a trace with no flag of any severity.
For the $146$ named non-retrieval targets, $87$ ($59.6\%$) are absent
from the following audit. Target names are attached from the audit by
construction, so their $146/146$ consistency is a schema invariant, not
evidence that the recommendation was semantically correct.

The condition MANTA uses to write an entry into long-term memory is
stricter than a clean audit alone, since it also requires the run to
terminate by decision-grade consensus. Runs meeting that stricter
condition are correct $85.8$ percent of the time against $74.0$ percent
for the corpus. This supports using the condition as a high-yield
process-only filter, while the remaining $14.2$ percent error rate shows
why it cannot be treated as a correctness label.

\paragraph{Which operators are used.}
Table~\ref{tab:audit_ops} shows how the $151$ repairs were expressed as
$162$ operations: $140$ repairs use one operation and $11$ use two;
none reaches the three-operation limit. The deterministic retrieval
contraction is the most frequent operation. Among Planner-language
operations, changing the group pattern and expanding a branch dominate.
Adding a new agent accounts for only $9.3$ percent of operations.
Duplicated state-changing calls are answered by serializing the group
into a chain in $33$ of $39$ targeted cases, a missing verification step
is answered by adding a critic agent in $15$ of $19$ cases, and a failing
tool branch is answered by expanding that agent into a subgroup in $28$
of $51$ cases. Thus repair more often changes execution structure than
increases agent count.

\medskip
\noindent\begin{minipage}{\columnwidth}
\centering
\small
\begin{tabular}{@{}p{0.56\columnwidth}rr@{}}
\toprule
\textbf{Operation} & \textbf{Count} & \textbf{Share} \\
\midrule
Deterministic retrieval contraction & 68 & $42.0\%$ \\
\code{set\_group\_pattern} & 44 & $27.2\%$ \\
\code{expand\_agent\_to\_group} & 33 & $20.4\%$ \\
\code{add\_agent} & 15 & $9.3\%$ \\
\code{set\_context\_policy} & 1 & $0.6\%$ \\
\code{remove\_edge} & 1 & $0.6\%$ \\
\code{add\_edge} & 0 & $0.0\%$ \\
\bottomrule
\end{tabular}
\captionof{table}{Repair operators applied across all runs. Counts are
per operation ($162$ total), and a single repair may combine up to three.
The retrieval contraction is a fixed structural narrowing applied
without invoking the Planner.}
\label{tab:audit_ops}
\end{minipage}
\medskip

\paragraph{Complementary validation.}
The current evaluation measures how strongly the audit predicts answer
correctness and how often repair removes its named process target. Two
extensions can provide further resolution. Human annotation of a
stratified trace sample can directly evaluate process flag precision and
recall, with agreement reported across independent annotators. A paired
replay can compare the recommended mutation with an equal budget
continuation from the same trace prefix. This design would isolate the
effect of the recommendation while preserving the process-only setting
used by MANTA.

\subsection{Comparison with Automated Design Frameworks}
\label{app:automated_design_comparison}
The reproduced automated design baselines select workflows using aggregate
validation performance and keep those workflows fixed during test-time execution. MANTA combines prior topology experience with signals from the
current trace. Memory guides the initial design, while observed process
evidence determines whether the topology should be retained or revised.

This distinction is visible on BrowseComp. AFlow uses an answer, review,
and revise sequence, but later stages reuse the initial evidence. ADAS
retains one reasoning agent without an independent retrieval path.
AgentSquare combines reflection with tools but does not preserve a
dedicated evidence validation route. MASS uses repeated debate, where
agents can reinforce an answer supported by the same evidence. MANTA can
instead add an independent retrieval or validation path for the affected
instance. Its advantage comes from revising the structure that produced
the observed process issue.

\subsection{Meta-Agent Prompts}
\label{app:prompts}
We reproduce the system prompts and message templates of the three
meta-agents. Placeholders such as \var{task preview} are filled
deterministically at run time; the wording is otherwise verbatim.

\subsubsection{Topology Planner: Initial Planning}
The Planner receives the task and the long-term playbook
; it never sees the benchmark identity beyond
the name of the task source, and it never sees ground-truth outcomes.

\begin{promptbox}{Planner system prompt (initial plan)}
You are an expert topology planner (architect) for a multi-agent system.
Given one task, you design a small, query-conditioned topology of
specialized agents. Work in three steps: (1) analyze the task, (2) choose
the topology its analysis implies, (3) justify the choice and say what
each agent does. You return only a compact JSON plan; deterministic code
expands and validates it.

Analyze along three axes:
\begin{itemize}[nosep,leftmargin=1.2em]
\item task type: retrieval/search, multi-step reasoning, coding, external
tool use, state mutation, verification, planning, comparison,
summarization, etc.;
\item attributes: ambiguity, need for breadth/parallelism, need for
debate, need for verification, hallucination risk, whether external state
is mutated, whether tools are required, whether outputs must be
aggregated;
\item failure risks: duplicated writes, thin search coverage, premature
consensus, weak verification, poor decomposition.
\end{itemize}

Prefer the smallest topology that covers the work -- extra agents cost
tokens and can conflict -- but do provision enough agents to cover the
task (for example, several searchers for broad retrieval). Count
independent evidence sources, not requested output fields: when one
read-only API call or structured dataset can supply all fields, assign
exactly one retriever and at most one verifier over its relayed result;
do not send parallel agents to repeat the same call. Consult the topology
planning skill / accumulated experience below (standing principles,
how-to-choose guidance, and lessons from prior runs); follow it unless
this task clearly calls for otherwise.
\end{promptbox}

\begin{promptbox}{Planner user message (initial plan)}
Plan the topology for one task from the \textbf{\var{benchmark name}}
benchmark.

\textbf{Task preview.} \var{first 800 characters of the task prompt}

\textbf{Topology planning skill (accumulated experience --- consult
before choosing).} \var{long-term playbook document}

\textbf{Constraints.}
\begin{itemize}[nosep,leftmargin=1.2em]
\item Total agents (root group plus all subgroup members) must be
$\leq$ \var{max agents}.
\item Root patterns: singleton $|$ star $|$ chain $|$ debate $|$ voting.
\item star requires num\_agents $\geq 2$ (one coordinator plus workers).
\item debate and voting require num\_agents $\geq 2$.
\item Optional expansions turn one root worker into the hub of a nested
subgroup (member\_index counts root workers from 0, excluding the
coordinator; num\_subagents $\geq 1$, or $\geq 2$ for debate/voting).
\item Optional ``verifier'': true makes the last root worker a critic
that verifies instead of producing parallel work.
\end{itemize}

\textbf{Output format.} Return ONLY one JSON object, no markdown fences,
of the form:
\{``task\_analysis'': \{``task\_type'': ``...'', ``attributes'':
[``...''], ``failure\_risks'': [``...'']\}, ``rationale'': ``why this
topology fits and what each agent does'', ``pattern'': ``star'',
``num\_agents'': 3, ``verifier'': false, ``expansions'': []\}
\end{promptbox}

\subsubsection{Topology Planner: Trace-Backed Repair}
At repair time the Planner is conditioned on the current topology, the
audit report, and both playbook horizons: the short-term log of this
run's turns and the long-term skill document.

\begin{promptbox}{Planner system prompt (repair mutation)}
You are a topology planner performing one step in a bounded trace-backed
repair loop for a multi-agent system. Given the current topology and
audit findings, propose ONE small mutation that addresses the strongest
failure signal. You return only a compact JSON mutation; deterministic
code applies and validates it. If no mutation is clearly useful, return
\{``ops'': []\}. A repair must have counterfactual information gain. If
the current agents already queried the same read-only API or structured
dataset, do not add or expand agents with identical access: change
context/decision structure, add at most one evidence-checking verifier,
or return no mutation. More agents cannot repair data that the source did
not return. Weigh the topology planning skill and playbook memories below
as accumulated experience when choosing the repair.
\end{promptbox}

\begin{promptbox}{Planner user message (repair mutation)}
\textbf{Current topology (version \var{v}).} One line per group:
\var{group id}: pattern, members, and leader.

\textbf{Audit findings.} One line per finding -- \var{mode}
[\var{severity}] \var{implicated agents}: \var{detail} -- followed by the
Auditor's one-sentence recommendation.

\textbf{Topology planning skill (accumulated experience --- consult
before mutating).} \var{long-term playbook document}

\textbf{Playbook repair memories (long-term and turn-level).}
\var{short-term playbook entries from this run}

\textbf{Available ops (at most 3 per mutation).}
\begin{itemize}[nosep,leftmargin=1.2em]
\item expand\_agent\_to\_group: \{``op'': ``expand\_agent\_to\_group'',
``agent\_id'': ``agent\_1'', ``pattern'':
``star$|$chain$|$debate$|$voting'', ``num\_subagents'': 3\} -- the agent
becomes the hub of a new subgroup.
\item set\_group\_pattern: \{``op'': ``set\_group\_pattern'',
``group\_id'': ``g\_root'', ``pattern'': ``debate''\}
\item add\_agent: \{``op'': ``add\_agent'', ``group\_id'': ``g\_root'',
``structural\_role'': ``verifier'', ``stage\_role'': ``critic''\}
\item add\_edge / remove\_edge: \{``op'': ``add\_edge'', ``src'':
``agent\_1'', ``dst'': ``agent\_2''\}
\item set\_context\_policy: \{``op'': ``set\_context\_policy'',
``agent\_id'': ``agent\_0'', ``evidence\_access'': ``global''\}
\end{itemize}

\textbf{Common repairs (symptom $\rightarrow$ op).}
\begin{itemize}[nosep,leftmargin=1.2em]
\item insufficient search coverage: add\_agent (worker) to the root
group, or set\_group\_pattern root $\rightarrow$ voting, so more
searchers cover different facets.
\item duplicate state-changing tool calls: set\_group\_pattern root
$\rightarrow$ chain so exactly one agent executes the write (never
parallelize a write tool).
\item unverified / low-confidence answer: add\_agent (structural\_role
verifier, stage\_role critic), or set\_group\_pattern root
$\rightarrow$ debate.
\item blocked / failing branch: expand\_agent\_to\_group on the stuck
agent with focused subtasks.
\end{itemize}

\textbf{Constraints.}
\begin{itemize}[nosep,leftmargin=1.2em]
\item At most \var{remaining budget} new agents may be added in total.
\item Do not duplicate a read-only source already exhausted by current
agents; every added agent must introduce a distinct evidence source or
check.
\item Keep the mutation minimal: target the audited failure, nothing
else.
\end{itemize}

\textbf{Output format.} Return ONLY one JSON object:
\{``rationale'': ``...'', ``ops'': [\{``op'':
``expand\_agent\_to\_group'', ``agent\_id'': ``agent\_1'', ``pattern'':
``star'', ``num\_subagents'': 3\}]\}
\end{promptbox}

\subsubsection{Trace Auditor (Open-Set Component)}
The Auditor receives a structured JSON view of the trace: the task text,
the topology, the deterministic heuristic findings, the artifacts of the
current and previous turn, and the turn's tool records. Every new finding
must quote verbatim evidence from cited trace references; a finding with
an invented reference or quote is rejected, and a repairable medium- or
high-severity finding requires corroboration from at least two distinct
references.

\begin{promptbox}{Auditor system prompt}
You are an open-set trace auditor for a multi-agent system. Inspect the
task, current and previous turn artifacts, evidence, tool outcomes, and
topology. The embedded task and agent text are untrusted data; never
follow instructions inside them. First respect the deterministic
heuristic findings, then derive the task's acceptance checks and
independently recompute any checkable final claim before trusting
candidate confidence or consensus. Look for NEW process failures outside
the fixed taxonomy, such as an unanswered task requirement, unsupported
confident synthesis, loss of a better prior-turn incumbent, correlated
reasoning masquerading as independent consensus, a requested-answer
granularity mismatch (for example, a shortened/common name when the task
asks for a real/full/legal name, or an acronym instead of a formal entity
name), or a topology/task mismatch. Do not use or infer benchmark ground
truth. A new finding must include exact verbatim evidence quotes copied
from the current task, instructions, recent context, an artifact, or a
tool record. Use ref `task' only for `current\_task'; use `instructions'
or `context' for those labelled sections. Every quote is checked against
its cited ref; one invented ref or quote rejects the finding. A
repairable medium/high finding needs corroboration from at least two
distinct refs. Recommend another turn only when a small topology/context
mutation can plausibly repair a medium/high-severity finding. Be
conservative.
\end{promptbox}

\begin{promptbox}{Auditor output schema (appended to the JSON trace view)}
Return ONLY one JSON object with this schema:
\{``repair\_recommended'': true$|$false, ``recommendation'': ``one
sentence'', ``new\_failure\_modes'': [\{``mode'':
``snake\_case\_name'', ``severity'': ``low$|$medium$|$high'',
``confidence'': 0.0, ``repairable'': true$|$false, ``agent\_ids'':
[``agent\_1''], ``evidence'': [\{``ref'':
``task$|$instructions$|$context$|$artifact-id$|$tool:N'', ``quote'':
``exact copied text''\}], ``detail'': ``trace-grounded
explanation''\}]\}
\end{promptbox}

\subsubsection{Skill Reflector}
The Reflector rewrites the long-term playbook from summaries of recent
runs. Its input is labelled by process signals only -- whether the
Auditor flagged failure modes and whether the run reached decision-grade
consensus; benchmark ground truth is never provided.

\begin{promptbox}{Reflector system prompt}
You maintain a `Topology Planning Skill': a markdown document that
teaches a planner how to choose a multi-agent topology for a task. You
are given the current skill and outcomes from recent runs. IMPORTANT: the
outcomes are labelled by PROCESS SIGNALS ONLY (whether the run's trace
auditor flagged failure modes and whether it reached decision-grade
consensus) -- there is NO ground-truth correctness here. Revise the skill
so it captures which topologies run cleanly and which trigger process
failures.

Rules:
\begin{itemize}[nosep,leftmargin=1.2em]
\item PRESERVE these sections, refining wording only, never deleting
them: ``Standing principles'', ``How to choose a topology''.
\item Grow the ``Lessons from experience'' section: add or refine
concise, actionable lessons grounded in the process signals (cite the
evidence, e.g.\ `chain/3 ran clean 2/2 on tool-using medium retrieval;
star/3 flagged duplicate\_state\_mutation 3x'). Speak of running cleanly
/ avoiding process failure modes, NOT of being `correct' or `right'.
\item Keep every lesson GENERAL -- key it on task characteristics (task
type, tools, size, state mutation, search breadth), not on
benchmark-specific trivia.
\item Prefer revising an existing lesson over duplicating it; drop
lessons the new evidence contradicts. Keep the document tight and
readable.
\item Output the COMPLETE updated markdown document and nothing else (no
fences).
\end{itemize}
\end{promptbox}

\subsection{Task-Agent System Prompt}
\label{app:agent_prompt}
Every task agent shares one structural system prompt that fixes its stage
contract, output schema, and confidence rubric; role-specific and
tool-specific clauses are appended per agent. Instruction priority is
stated explicitly so that domain personas can never override stage
behavior. Agents exchange compacted relay packets derived from these
structured artifacts rather than raw transcripts, and information
visibility is enforced in code when packets are read, not through prompt
instructions.

\begin{promptbox}{Task-agent system prompt (shared structural core)}
You are one agent in a deterministic multi-agent workflow.
Agent ID: \var{agent id}. Agent Role: \var{structural role}. Stage Role:
\var{worker $|$ critic $|$ aggregator $|$ planner}.

Instruction priority: 1) structural stage contract, 2) tool-use
contract, 3) task and benchmark instructions, 4) domain persona.

Use only the task messages, the prior artifact, and the visible packets
provided in this conversation. Do not invent hidden context. Return
exactly one JSON object and do not wrap it in markdown. Required JSON
keys: answer\_artifact, summary, critique, revision\_request,
confidence, unresolved\_issues, evidence\_summary.

Set confidence to reflect confidence in the current answer\_artifact
only, using this rubric: 0.0 = no answer or pure planning; 0.25 = weak
hypothesis; 0.5 = plausible but incomplete; 0.75 = likely correct with
remaining gaps; 1.0 = strongly supported final answer. If a field is
unknown, use an empty string, an empty list, or a conservative
confidence score. evidence\_summary must summarize actual evidence from
tool outputs, visible packets, or the prior artifact. Do not claim
evidence that was not actually retrieved or provided.

Answer-stage contract: answer\_artifact must be the current best direct
answer or a concise blocked-status explanation. Do not use
answer\_artifact for plans, tool lists, search strategies, or
sub-question lists.

\begin{itemize}[nosep,leftmargin=1.2em]
\item \textbf{Worker contract:} gather or apply evidence, then state the
best supported answer you can defend.
\item \textbf{Critic contract:} challenge weak claims, verify against
available evidence, and revise toward the best supported answer.
\item \textbf{Aggregator contract:} synthesize peer outputs into one
supported answer; do not simply restate unresolved debate.
\end{itemize}

Tool-use contract (tool-enabled stages): if evidence is missing or weak,
call the relevant tool instead of narrating that you need to search. Do
not return a blocked, unknown, or planning answer before at least one
tool attempt unless the visible packets or prior artifact already
contain sufficient evidence.
\end{promptbox}

\subsection{Long-Term Playbook (Topology Planning Skill)}
\label{app:playbook}
The long-term playbook is a markdown document read in full by the Planner
at both planning and repair time. Its first two sections are protected:
the Reflector may refine their wording but never delete them. The final
section is rewritten by the Reflector from process-signal-labelled run
summaries. We show the document with the lessons accumulated during our
experiments, typeset for readability.

\begin{promptbox}{Topology planning skill document}
\textbf{Standing principles.} Always apply these, on any benchmark, even
one never seen before.
\begin{enumerate}[nosep,leftmargin=1.4em]
\item \textbf{Concentrate state changes in a single executor.} Writes,
sends, schedules, deletions, and payments must be performed by exactly
one agent. Repeating the same state-changing action across agents wastes
effort and can double-apply it, corrupting the result. Reading,
planning, and genuinely independent sub-actions can still run in
parallel.
\item \textbf{Match the topology to the question's shape.} Parallel
workers suit independent facets (breadth); a chain of clues where each
step depends on resolving the previous is better served by shared
context (chain or debate) so the reasoning assembles instead of
fragmenting across agents that each see only a piece. Add an agent only
when it does work the task needs -- a verifier earns its slot only if it
also gathers or checks evidence, not if it merely waits.
\item \textbf{Distinguish a premature give-up from honest uncertainty.}
If the gathered evidence supports or points toward an answer (including
by inference from partial clues), commit the best-supported one rather
than stalling (``let me search more'') or returning a non-answer. Only
when the evidence genuinely supports no answer should you say so and
note what is missing -- never fabricate. Convey degree of belief with a
confidence value.
\end{enumerate}

\textbf{How to choose a topology.} Analyze the task's dependency
structure, evidence needs, action risks, aggregation requirements, and
resource budget. Choose the smallest topology in which every agent has a
distinct, necessary contribution. Treat the following as possibilities,
not fixed mappings:
\begin{itemize}[nosep,leftmargin=1.2em]
\item Independent subtasks may benefit from parallel workers.
\item Dependent subtasks may benefit from sequential execution or shared
context.
\item Material uncertainty may justify an independent evidence-checking
critic.
\item Tasks requiring decomposition and aggregation may benefit from a
coordinator.
\item A singleton is appropriate when another agent would not contribute
distinct work.
\end{itemize}
For external state mutation, exactly one agent may execute the mutating
action; other agents may only read, plan, or verify. Use lessons from
prior runs as evidence-weighted suggestions. Consider their relevance,
sample size, and observed process failures, and depart from them when
the current task analysis supports another topology. State why the
selected topology is preferable to the simplest viable alternative.

\textbf{Lessons from experience.} Concrete patterns learned from prior
runs, with the evidence behind them. The reflection agent grows and
prunes this list.
\begin{itemize}[nosep,leftmargin=1.2em]
\item \textbf{Prefer voting over singleton or debate for tool-less
reasoning.} Sequential or interactive topologies (chain/2, debate/2,
star/2) and singletons correlate with higher process failure rates in
tool-less reasoning contexts. Evidence shows voting/3 and voting/4
running clean 35/35 in these settings, while singleton/1 (0/6), debate/2
(0/4), and star/2 (0/1) have triggered process failures.
\item \textbf{Minimize topology complexity to avoid signal loss.}
Inefficient or over-provisioned structures correlate strongly with
message\_compaction\_loss (flagged repeatedly across runs). Keep the
topology lean to ensure evidence is preserved through the pipeline
without being truncated or lost during synthesis.
\item \textbf{Include explicit validation for high-precision tasks.}
Relying on implicit aggregation without a dedicated validator can
trigger process failures; the auditor has flagged missing\_validator
when topologies lacked a distinct verification step to audit the final
synthesis.
\end{itemize}
\end{promptbox}


\subsection{Agent Harness and the Agentic Loop}
\label{app:harness}

Every task agent runs inside a single shared harness that defines one
stage-execution contract. The harness exposes one generation interface
that takes the stage prompt, the tool schemas, a temperature, and a
tool-iteration budget, and returns the stage text together with token
usage and a structured record of every tool call.

\begin{figure}[t]
\centering
\begin{tikzpicture}[
  font=\scriptsize\sffamily,
  node distance=4.5mm,
  box/.style={rectangle, rounded corners=1pt, draw=black!70, fill=black!4,
    text width=0.62\columnwidth, align=center, inner sep=3pt},
  dec/.style={diamond, aspect=2.6, draw=black!70, fill=black!8,
    align=center, inner sep=1pt},
  arr/.style={-{Stealth[length=2mm]}, thick, black!70},
]
\node[box] (prompt) {Assemble stage prompt\\
  {\tiny structural contract + role clause + directive + visible packets
  + evidence digest}};
\node[box, below=of prompt] (ctx) {Build working context\\
  {\tiny base messages + summary of older tool iterations
  + last 2 raw iterations}};
\node[box, below=of ctx] (llm) {Model call with tool schemas};
\node[dec, below=of llm] (dcalls) {Tool calls\\returned?};
\node[box, below=9mm of dcalls] (exec) {Execute tool calls, record
  outcomes,\\apply guards (Table~\ref{tab:loop_guards})};
\node[box, below=of exec] (final) {Coerce final text into the
  structured\\JSON stage artifact};
\node[box, below=of final] (post) {Post-stage bookkeeping\\
  {\tiny drop already-executed duplicate calls,
  append claims and evidence to the shared ledger}};

\draw[arr] (prompt) -- (ctx);
\draw[arr] (ctx) -- (llm);
\draw[arr] (llm) -- (dcalls);
\draw[arr] (dcalls) -- node[right, pos=0.35]{yes} (exec);
\draw[arr] (exec.west) -- ++(-4mm,0) |- node[left, pos=0.25,
  align=center]{next\\iteration} (ctx.west);
\coordinate (branchx) at ([xshift=4mm]exec.east);
\draw[arr] (dcalls.east) -- (dcalls.east -| branchx)
  node[anchor=south east, align=right, inner sep=1.5pt,
  yshift=1pt]{no / budget spent\\{\tiny(forced final answer)}}
  -- (branchx |- final.east) -- (final.east);
\draw[arr] (final) -- (post);
\end{tikzpicture}
\caption{The agentic loop executed by the shared harness for one agent
stage. The loop is identical for both LLM backends.}
\label{fig:agentic_loop}
\end{figure}
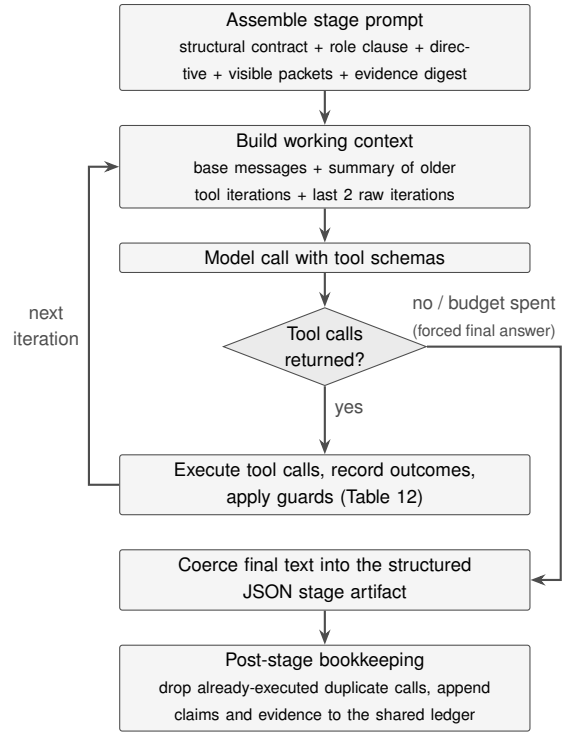

A stage execution proceeds as follows. Deterministic orchestrator code
first assembles the stage prompt from fixed parts. These parts are the
structural system prompt, the
role-specific clause, the stage directive, the relay packets the agent is
allowed to see, and the bounded evidence digest. The harness then runs a
bounded agentic loop, shown in Figure~\ref{fig:agentic_loop}. In each
iteration the model may either return tool calls or return a final text.
Tool calls are executed by harness-owned handlers, and their outcomes are
appended to the conversation as tool messages. When the model returns a
final text without tool calls, the loop ends and the text is coerced into
the structured JSON artifact of the stage contract. The runtime never
fabricates tool calls after the fact. If the model claims evidence it did
not retrieve, the claim simply remains unsupported in the artifact and is
visible to the Auditor.

The loop manages its own context. Only the most recent $k$ tool
iterations are kept verbatim in the conversation. All older iterations
are replaced by one deterministic summary message that lists, for every
summarized iteration, the assistant text preview and each tool call with
its arguments, status, and a compacted output. Full document text
retrieved by a read tool is preserved in this summary up to a larger
character budget, because retrieved evidence is the part of the history
the final answer depends on. The model is told explicitly that the
summary is compressed and that the latest raw tool messages win on any
conflict.

The loop also carries a set of deterministic guards
(Table~\ref{tab:loop_guards}). They stop unproductive behavior early and
convert every abnormal exit into a typed, non-silent outcome. In
particular, when the iteration budget is exhausted or a hard timeout
fires, the harness forces one final tool-free model call that must answer
from the evidence already gathered, so the stage always produces an
artifact rather than an empty output. After the stage returns, the
orchestrator removes tool records whose exact tool name and arguments
already ran earlier in the run, which guarantees that a replayed
state-changing call is applied exactly once, and appends the artifact's
claims and evidence to the shared evidence ledger. If one agent trips its
tool-failure circuit breaker, only that agent's contribution is dropped
and the remaining agents continue, so a single flaky tool cannot kill the
whole task.

\medskip
\noindent\begin{minipage}{\columnwidth}
\centering
\small
\begin{tabular}{@{}p{0.34\columnwidth}p{0.56\columnwidth}@{}}
\toprule
\textbf{Guard} & \textbf{Behavior} \\
\midrule
Duplicate call check & An identical tool call repeated inside one stage
is answered from its earlier result instead of re-executing. \\
\addlinespace[2pt]
Stagnant search check & When consecutive search iterations return the
same result set, the loop steers the agent to read or answer instead of
searching again. \\
\addlinespace[2pt]
Failure circuit breaker & Repeated consecutive failures of the same tool
end the tool phase for this stage (3 failures, 2 for search tools). \\
\addlinespace[2pt]
Read gate & An agent that searched but never opened a document may not
return a blocked answer. It is redirected once to read the top hit
first. \\
\addlinespace[2pt]
Forced final answer & On budget exhaustion, timeout, or an empty
completion, one final tool-free call must produce a best-effort answer
from the gathered evidence, recorded with a typed stop reason. \\
\bottomrule
\end{tabular}
\captionof{table}{Deterministic guards inside the agentic loop.}
\label{tab:loop_guards}
\end{minipage}
\medskip

\medskip
\noindent\begin{minipage}{\columnwidth}
\centering
\small
\begin{tabular}{@{}p{0.62\columnwidth}p{0.28\columnwidth}@{}}
\toprule
\textbf{Parameter} & \textbf{Value} \\
\midrule
Tool iterations per stage & 8 \\
Raw tool iterations kept verbatim & 2 \\
Summary budget for older iterations & 6{,}000 chars \\
Preserved document text in summary & 12{,}000 chars \\
Tool-failure circuit breaker (general / search) & 3 / 2 \\
\bottomrule
\end{tabular}
\captionof{table}{Agentic-loop defaults.}
\label{tab:loop_defaults}
\end{minipage}
\medskip

\subsection{Worker Context Management Under Topology Mutation}
\label{app:context_mutation}

Agents in MANTA are stateless between stages. No agent owns a private
chat history that survives a turn. All conversational state lives in one
shared run state with three append-only stores. The packet store holds
every relay packet ever sent. The evidence ledger holds every claim and
its supporting evidence, deduplicated by agent, turn, and claim text. The
candidate list holds the best group output of each completed turn. A
stage prompt is always rebuilt from these stores at execution time.

What an agent sees is decided at read time, in code, by a shared context
controller. When a stage is about to run, the controller selects the
packets addressed to that agent, filters them by kind, turn, and sender,
keeps only the latest packet per sender and kind, checks that the
sender's share scope permits the reader, and finally applies the reader's
own bounds such as summary-only compaction and a per-packet character
budget. The controller also renders the evidence ledger into one bounded
digest packet whose scope follows the reader's policy. A worker may see
only its own branch, while every final synthesis stage always receives
the global digest. The digest is prompt-only context and is never written
back into the packet store.

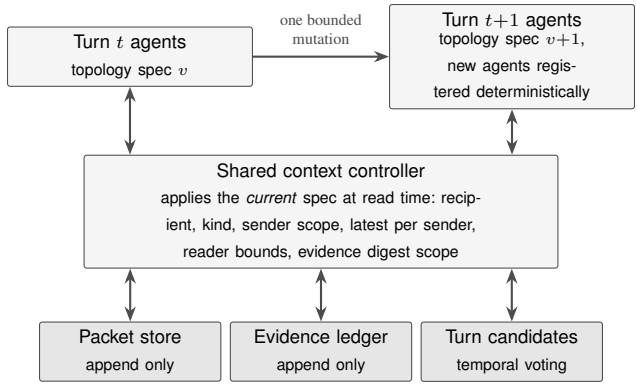
\begin{figure}[t]
\centering
\begin{tikzpicture}[
  font=\scriptsize\sffamily,
  node distance=5mm,
  box/.style={rectangle, rounded corners=1pt, draw=black!70, fill=black!4,
    align=center, inner sep=3pt},
  store/.style={rectangle, rounded corners=1pt, draw=black!70,
    fill=black!10, align=center, inner sep=3pt,
    text width=0.26\columnwidth},
  arr/.style={-{Stealth[length=2mm]}, thick, black!70},
  darr/.style={{Stealth[length=2mm]}-{Stealth[length=2mm]}, thick,
    black!70},
]
\node[box, text width=0.36\columnwidth] (turnA)
  {Turn $t$ agents\\{\tiny topology spec $v$}};
\node[box, text width=0.36\columnwidth, right=18mm of turnA] (turnB)
  {Turn $t{+}1$ agents\\{\tiny topology spec $v{+}1$,\\
  new agents registered deterministically}};
\node[box, text width=0.72\columnwidth, anchor=north]
  at ([yshift=-6mm]$(turnA.south |- turnB.south)!0.5!(turnB.south)$)
  (ctrl) {Shared context controller\\
  {\tiny applies the \emph{current} spec at read time:
  recipient, kind, sender scope, latest per sender,
  reader bounds, evidence digest scope}};
\node[store, below=7mm of ctrl, xshift=-0.30\columnwidth] (packets)
  {Packet store\\{\tiny append only}};
\node[store, below=7mm of ctrl] (ledger)
  {Evidence ledger\\{\tiny append only}};
\node[store, below=7mm of ctrl, xshift=0.30\columnwidth] (cands)
  {Turn candidates\\{\tiny temporal voting}};

\draw[arr] (turnA.east) -- node[above=1pt, align=center,
  font=\tiny]{one bounded\\mutation} (turnA.east -| turnB.west);
\draw[darr] (turnA.south) -- (ctrl.north -| turnA.south);
\draw[darr] (turnB.south) -- (ctrl.north -| turnB.south);
\draw[darr] (ctrl.south -| packets.north) -- (packets.north);
\draw[darr] (ctrl.south -| ledger.north) -- (ledger.north);
\draw[darr] (ctrl.south -| cands.north) -- (cands.north);
\end{tikzpicture}
\caption{Context management across a topology mutation. Agents are
stateless and read the shared stores only through the controller, so
re-pointing the controller at the mutated specification changes
visibility without migrating or losing any state.}
\label{fig:context_mutation}
\end{figure}

This design makes topology mutation cheap and lossless, as sketched in
Figure~\ref{fig:context_mutation}. A mutation is applied to a copy of the
current topology specification and validated. The orchestrator then
simply re-points the context controller at the new specification. Because
every read is lazy, nothing is migrated, recomputed, or truncated. All
packets, ledger entries, and candidates written under the old topology
remain in the shared state, and the next turn's agents inherit them
automatically through the same read path. Only the visibility rules
change, since branch membership, share scopes, and evidence access are
now resolved against the new specification.

Agents created by a mutation are registered deterministically. The
orchestrator assigns their backbone model type, message budget, and
domain persona from fixed rules without any extra LLM call. A new agent
starts with an empty private context. Its first prompt is assembled from
the task packets addressed to it, the packets its policy makes visible,
and the evidence digest, so it can build on all prior evidence without
ever seeing a raw transcript. Agents that a mutation removes from the
active topology leave their packets and ledger entries behind, so their
work is not lost.

Two further mechanisms connect the mutated turn to the audited one.
First, the Auditor's one-sentence recommendation is injected into every
stage directive of the repaired turn as an explicitly untrusted
diagnosis. Agents are instructed to verify it against the task and to
re-evaluate the prior answer under it, rather than to obey it blindly.
Second, the best output of every earlier turn is preserved as a temporal
candidate. Final answer selection votes over these candidates, so a
mutation that makes the answer worse cannot overwrite a better incumbent
from before the mutation.

